\documentclass{ieeeaccess}

\usepackage{cite}
\usepackage{amsmath,amssymb,amsfonts}
\usepackage{graphicx}
\usepackage{textcomp}
\def\BibTeX{{\rm B\kern-.05em{\sc i\kern-.025em b}\kern-.08em
    T\kern-.1667em\lower.7ex\hbox{E}\kern-.125emX}}

\usepackage{booktabs, multirow} 
\usepackage{tabularx}

\usepackage{amsmath}
\usepackage{colortbl}
\usepackage{flushend}

\usepackage{algorithm}
\usepackage{algpseudocode}
\usepackage{color,soul}

\usepackage{mathtools}
\AtBeginEnvironment{tcolorbox}{\small}
\usepackage{hyperref}
\usepackage{orcidlink}

\usepackage[group-separator={,}]{siunitx}
\usepackage{rotating}
\usepackage{tikz}
\usetikzlibrary{shapes.geometric}
\NewSpotColorSpace{PANTONE}
  \AddSpotColor{PANTONE} {PANTONE3015C} {PANTONE\SpotSpace 3015\SpotSpace C} {1 0.3 0 0.2}
  \SetPageColorSpace{PANTONE}%
\makeatletter

\usepackage{fontawesome}
\newcommand\mathcircled[1]{%
  \mathpalette\@mathcircled{#1}%
}
\newcommand\@mathcircled[2]{%
  \tikz[baseline=(math.base)] \node[draw,circle,inner sep=3pt] (math) {$\m@th#1#2$};%
}

\usepackage{siunitx}
\tikzset{
    buffer/.style={
        draw,
        shape border rotate=-30,
        isosceles triangle,
        isosceles triangle apex angle=85,
        fill=black,
        node distance=0.1em,
        scale=0.1
    }
}

\begin{document}
\history{Date of publication xxxx 00, 0000, date of current version xxxx 00, 0000.}
\doi{10.1109/ACCESS.2017.DOI}

\title{Metaheuristics and Large Language Models Join Forces: Towards an Integrated Optimization Approach}

\author{\uppercase{Camilo Chacón Sartori}~\orcidlink{0000-0002-8543-9893}\authorrefmark{1},
\uppercase{Christian Blum~\orcidlink{0000-0002-1736-3559}\authorrefmark{1}, Filippo Bistaffa~\orcidlink{0000-0003-1658-6125}\authorrefmark{1}, Guillem Rodríguez Corominas~\orcidlink{0000-0002-3863-2017}}\authorrefmark{1}}
\address[1]{Artificial Intelligence Research Institute (IIIA-CSIC), Bellaterra,08193, Barcelona, Spain}

\tfootnote{The research presented in this paper was supported by grants TED2021-129319B-I00 and PID2022-136787NB-I00 funded by MCIN/AEI/10.13039/ 501100011033.}

\corresp{Corresponding author: Camilo Chacón Sartori (e-mail: cchacon@iiia.csic.es).}

\begin{abstract}
Since the rise of Large Language Models (LLMs) a couple of years ago, researchers in metaheuristics (MHs) have wondered how to use their power in a beneficial way within their algorithms. This paper introduces a novel approach that leverages LLMs as pattern recognition tools to improve MHs. The resulting hybrid method, tested in the context of a social network-based combinatorial optimization problem, outperforms existing state-of-the-art approaches that combine machine learning with MHs regarding the obtained solution quality. By carefully designing prompts, we demonstrate that the output obtained from LLMs can be used as problem knowledge, leading to improved results. Lastly, we acknowledge LLMs' potential drawbacks and limitations and consider it essential to examine them to advance this type of research further. Our method can be reproduced using a tool available at: \url{https://github.com/camilochs/optipattern}.
\end{abstract}


\titlepgskip=-15pt

\maketitle

\section{Introduction}

The advent of Large Language Models (LLMs) has altered the Natural Language Processing (NLP) landscape, empowering professionals across diverse disciplines with their remarkable ability to generate human-like text. Models like OpenAI's GPT~\cite{openai2023gpt4}, Meta's Llama~\cite{touvron2023llama}, and Anthropic's Claude 3~\cite{Anthropic2024-ee} have become indispensable collaborators in many peoples' daily lives; giving rise to innovative products such as ChatGPT for general use, GitHub Copilot for code generation, DALL-E 2 for image creation, and a multitude of voice generators, including OpenAI's text-to-speech API and ElevenLabs's Generative Voice AI. Currently, LLMs are being experimentally applied across various fields, yielding mixed results \cite{ALMEIDA2024104145}. While some applications seem questionable, others exhibit spectacular outcomes. One of the most contentious applications is using LLMs for tasks necessitating mathematical reasoning. Given LLMs' inherently probabilistic nature, this application was once deemed implausible. However, recent findings suggest a shift in perspective, particularly with LLMs boasting vast parameter counts~\cite{ahn2024large}. As LLMs continue to scale, new capabilities emerge~\cite{wei2022emergent}. Crucially, these opportunities are contingent upon the thoughtful design of prompts, which helps mitigate the risk of LLMs providing irrelevant or inaccurate responses~\cite{wan2024efficient}. 

Whenever a new technology emerges, it is natural to wonder if it might enhance an existing one. In combinatorial optimization, metaheuristics (MHs)~\cite{10.5555/1941310} have been established as effective approximate algorithms for tackling complex, NP-hard problems. While they excel in rapidly providing good-enough solutions, they depend heavily on domain-specific knowledge. To address this limitation, researchers have explored the integration of MHs with other approaches, including exact algorithms and Machine Learning (ML). While combining MHs with exact algorithms has shown promise~\cite{BLUM20114135}, a successful integration demands significant technical expertise. Alternatively, incorporating ML techniques within MHs can provide valuable problem insights~\cite{KARIMIMAMAGHAN2022393}, but these approaches often require specialized know-how or, in the case of Deep Learning (DL), substantial datasets and computational resources for training~\cite{barocas2023fairness}. This paper seeks another direction. We delve into the potential of LLMs to create a novel hybrid approach that combines the strengths of MHs and LLMs. 
\subsection{Our contribution}


We present a novel approach to enhance MHs performance utilizing LLMs' output. Our method diverges from existing techniques in two key aspects:

\begin{itemize}
    \item Rather than employing LLMs to generate MHs (e.g., ~\cite{10.1145/3583133.3596401, Liu2023-bq, ma2024llamoco}), our approach leverages them as pattern recognition tools for problem instance metrics. This strategy allows for seamless integration into existing MHs by introducing an additional LLM-provided parameter.
    \item Unlike methods that use LLMs as direct optimizers for natural language-described problems~\cite{yang2024large, guo2024optimizinglargelanguagemodels, ma2024largelanguagemodelsgood}---a technique limited to simple optimization tasks due to LLMs' stochastic nature---our approach tackles complex optimization challenges by using LLMs to identify and track pertinent information within the problem instance.
\end{itemize}

This dual-faceted approach represents a significant advancement in the integration of LLMs with metaheuristic optimization techniques, offering a more robust and versatile framework for tackling a wide array of optimization challenges. Thus, we employ LLMs not as an oracle providing final answers (i.e., it is not an \emph{end-to-end} approach according to the classification by Bengio et al.~\cite{bengio2020machine}) but as an \textit{intermediate step}, assisting in pattern detection within the metrics' values (see Figure~\ref{fig:integration}), i.e., it is an \emph{hybrid} one.

We validate our proposed \texttt{MH+LLM} integration using the \textit{Multi-Hop Influence Maximization in Social Networks} problem, demonstrating improved performance over the current state-of-the-art approach, which combines MH with deep learning (DL)~\cite{9909110}. Therefore, we believe this approach can unlock new possibilities for improving MHs by leveraging generative AI to tackle complex pattern recognition tasks. 

The paper is organized as follows. Section~\ref{sec:background} examines existing approaches for integrating ML into MHs and provides an overview of existing research on applying LLMs in optimization. Section~\ref{sec:problem} formally defines the NP-Hard combinatorial optimization problem that serves as an example for our study. Our proposed integration strategy for combining MH and LLMs is presented in Section~\ref{sec:integration}. The empirical evaluation of our hybrid approach is detailed in Section~\ref{sec:empirical-evaluation}, where we introduce a comprehensive three-dimensional framework for assessment and provide a visual analysis of the algorithm's performance. Section~\ref{sec:discussion} identifies remaining open research questions and discusses the current limitations of LLMs. Finally, Section~\ref{sec:conclusion} concludes the paper by summarizing our key findings. Moreover, future research directions are mentioned.

\subsection{Reproducibility}\label{subsec:repository}

Recognizing the importance of reproducibility in our field~\cite{10.1145/3466624}, and the potential challenges introduced by new technologies, we have developed a tool called \texttt{OptiPattern} (\textit{LLM-Powered Pattern Recognition for Combinatorial Optimization}) that automates our hybridization method---detailed in Section~\ref{sec:integration}---to ensure greater ease and accuracy in replication. This tool allows researchers to input a problem instance, generating the full prompt in response. Furthermore, by incorporating an LLM API key, the tool can return node-specific probabilities, which can then be integrated into the metaheuristic.\footnote{\url{https://github.com/camilochs/optipattern}} This is essential, as replicating prompts can be complex and prone to errors.

\section{Background}\label{sec:background}
\subsection{Machine Learning for Enhancing Metaheuristics in Combinatorial Optimization}

Metaheuristics (MHs) are approximate algorithms that have proven effective in solving complex optimization problems, especially combinatorial optimization problems (COPs). COPs are characterized by discrete variables and a finite search space. Although MHs have been shown to deliver good results in reduced computation times, they do not guarantee finding the optimal solution. Moreover, their success often hinges on the availability of problem-specific knowledge. Each problem instance is treated similarly, applying problem-specific heuristics and (generally) relying on a stochastic behavior. Along these lines, the community aims to innovate by integrating techniques from various domains to improve MHs' performance and address the limitations of MH techniques. In particular, \emph{hybrid} approaches based on the combination with (1) exact algorithms~\cite{BLUM20114135} and (2) learning techniques~\cite{KARIMIMAMAGHAN2022393} have been explored. Currently, the primary focus has shifted towards the second option. Especially the integration of machine learning (ML) techniques has recently resulted in a multitude of different hybrid approaches. Researchers have explored various strategies to integrate ML into MHs~\cite{KARIMIMAMAGHAN2022393}. In particular, ML might be used for the following purposes in MHs: algorithm selection (determining the best MH for a given problem), fast approximate fitness evaluation in the presence of costly objective functions, initialization (generating high-quality initial solutions), and parameter configuration (optimizing the numerous parameters of an MH, which is crucial for its performance). These strategies utilize learning techniques to analyze numerous cases and scenarios, uncovering hidden patterns in the data. By identifying these patterns, MHs can extract general principles that apply to a broad range of situations. This enhances MH's decision-making process, increasing their performance and adaptability.\footnote{The reverse integration, which involves enhancing ML architectures with MH techniques, is beyond the scope of this study.} 

Beyond classical ML methods (supervised learning: logistic regression, decision trees, support vector machines; unsupervised learning: k-means clustering, principal component analysis), there are approaches from several research subfields within the discipline that have been leveraged and tailored for their use in MHs. Each subfield is characterized by its unique methods, strategies, and possibilities. For instance, deep learning (DL) and reinforcement learning (RL) are two areas that have proven particularly promising for their integration with MHs.

DL employs many-layered artificial neural networks to automatically learn complex data representations, whereas RL focuses on sequential decision-making to achieve good results using a trial-and-error learning process. Methods from both fields have found valuable applications in the realm of MHs, enhancing their ability to find high-quality solutions. In fact, a growing body of research demonstrates this hybrid approach's success. In the following, we provide short descriptions of exemplary hybridization approaches from three different categories:
\begin{itemize}
    \item\texttt{MH+ML}: In a study by Sun et al.~\cite{SUN2022105769}, ML techniques were incorporated into the metaheuristic Ant Colony Optimization (ACO) to address the orienteering problem. The authors improved the solution construction process of ACO by utilizing guided predictions based on engineered features. In~\cite{10.1007/978-3-030-53552-0_15}, the authors developed a novel algorithm that combines a metaheuristic with decision trees to address the classic vehicle routing problem.
    \item\texttt{MH+DL}: Examples of this type of hybridization can be found abundantly in the literature of recent years. For instance, in~\cite{9909110}, the authors presented a novel approach that uses a Graph Neural Network (GNN) for learning heuristic information that is then used by a Biased Random Key Genetic Algorithm (BRKGA) to translate random keys into solutions to the tackled problem. Another example concerns~\cite{liu2023machine} where the authors apply different GNN architectures for parametrizing the neighbor selection strategy in Tabu Search (TS) and in Large Neighborhood Search (LNS).
    \item\texttt{MH+RL}: RL has recently been used in numerous hybrid approaches. For instance, in~\cite{huber2021learning}, the authors devised a method for learning the heuristic function of beam search in the context of two variants of the Longest Common Subsequence (LCS) problems. In \cite{FENOY2024104064}, the authors proposed a hybrid approach comprised of an attention-based model trained with RL and combined with a more classical optimization method for the formation of collectives of agents in real-world scenarios, showing that it reaches the performance of state-of-the-art solutions while being more general. Furthermore, a variable neighborhood search (VNS) based on Q-learning was devised for a machine scheduling problem in~\cite{alicastro2021reinforcement}. Finally, we also mention~\cite{chaves2021adaptive}, where RL is used for adapting the parameters of a BRKGA during the evolutionary process. 
\end{itemize}

While these hybridization strategies offer potential solutions, they each come with their own set of drawbacks. For instance, the manual feature selection process in many \texttt{MH+ML} approaches relies heavily on the expertise of a specialist. Concerning \texttt{MH+DL} approaches, the system's generalization ability might be hindered by lacking a large and diverse enough dataset. As for \texttt{MH+RL} hybrids, the complexity lies in defining the action space, rewards, and learning policies in a clear and effective manner. Moreover, all three approaches share similar technical challenges: the complexity of replicating models, the time-consuming process of data collection and preparation, and the computational demands of generalization, especially for large-scale problems or complex optimization tasks~\cite{barocas2023fairness}. 

To explore innovative methods for enhancing the performance of MHs and considering the usefulness and potential of LLMs, which will be discussed in the following subsection, in this paper, we explore a novel hybrid approach: \texttt{MH+LLM} (see Section~\ref{sec:integration}). Utilizing the capabilities of LLMs---while being aware of their limitations---we aim to enhance the problem-solving capabilities of MHs and open up new avenues for tackling complex optimization problems.

\subsection{LLMs as Pattern Recognition Engines}\label{section:llm}

LLMs have breathed new life into the field of NLP. These high-level language models employ billions of parameters and exhibit an outstanding ability to learn from data. Models like GPT-4o (OpenAI)\footnote{\url{https://openai.com/index/hello-gpt-4o}.}, Claude-3-Opus (Anthropic)\footnote{\url{https://www.anthropic.com/news/claude-3-family}.}, Gemini 1.5 (Google)\footnote{\url{https://deepmind.google/technologies/gemini}.}, Mixtral 8x22b (Mistral AI)\footnote{\url{https://mistral.ai/news/mixtral-8x22b}.}, and Command-R+ (Cohere)\footnote{\url{https://docs.cohere.com/docs/command-r-plus}.}, as well as tools built on top of them---such as ChatGPT, GitHub Copilot, and Bing Chat---have demonstrated that we are in the presence of a groundbreaking technology. Unlike previous advancements, this new wave of AI is no longer limited to experts; instead, it is accessible to anyone who can grasp its benefits.

LLMs are generative AI models that produce text sequentially, predicting each token based on the previous ones. This is made possible by the Transformer, a groundbreaking DL architecture that revolutionized the field of NLP. Proposed by Vaswani et al.~\cite{vaswani2023attention}, it introduces the concept of self-attention, allowing the model to contextually select the most suitable words. The Transformer derives its name from its ability to \textit{transform} a set of vectors in a given representation space to a new set of vectors with identical dimensions but in a different space. By assigning varying weight values to each input, the attention mechanism leverages inductive biases related to sequential data~\cite{M2023-wp}. This architecture also takes advantage of the capabilities of high-performance hardware due to its parallelizable nature. As a result, the Transformer generates words that seamlessly fit the context---although it lacks factual verification---marking a significant advancement in NLP tasks.

LLMs have found uses in various domains, including the interpretation of complex results like chemical compounds or images~\cite{singh2024rethinking}, where they provide insights and explanations. Additionally, LLMs are being applied as autonomous agents that leverage external tools and resources to accomplish tasks~\cite{guo2024large}. Furthermore, these models demonstrate progress in domains once thought to be out of the reach of their capabilities, including mathematical reasoning and optimization tasks~\cite{yang2024large}. While there are still numerous obstacles to overcome~\cite{ahn2024large}, these advancements showcase the potential of these models to address intricate cognitive problems.

Recent research on leveraging LLMs for optimization has primarily explored two paths: first, formulating optimization problems within the prompt and requesting the LLM to solve the described problem, typically for straightforward optimization tasks~\cite{yang2024large, guo2024optimizinglargelanguagemodels, ma2024largelanguagemodelsgood}; and second, automating code generation to enhance optimization algorithms~\cite{10.1145/3583133.3596401, Liu2023-bq, ma2024llamoco}. 
While both strategies are effective in leveraging LLMs to address certain weaknesses of MHs, they fail to account for the significance of problem instances, as different instances can produce varying results in an MH. Our approach seeks to tackle this challenge, functioning as a complementary tool rather than a rival to current strategies. Thus, we present a novel integration that utilizes LLMs to enhance the effectiveness of metaheuristic search processes. Although a recent study has shown that LLMs can detect patterns across various tasks~\cite{mirchandani2023large}, no method has yet been developed that employs LLMs as \textit{pattern recognition engines} in combinatorial optimization problems. Acknowledging the challenges related to LLMs (discussed in the subsequent Section~\ref{obstacles-llm}), we identify numerous opportunities for progress in this area (see Section~\ref{sec:integration}).

\subsubsection{Obstacles and Opportunities}\label{obstacles-llm}

LLMs are an emerging technology that is still evolving. Despite demonstrating usefulness across various applications, as seen above, we believe that our research has mitigated two key risks associated with LLMs:

\begin{enumerate}
    \item Training these models requires an immense amount of diverse data, from social media posts to books, and an equally vast amount of computing power. As a result, the leaders in the LLM industry tend to be well-funded private companies with significant resources and high valuations. This creates a high barrier to entry for startups or under-funded research centers, which often lack the necessary infrastructure to compete directly with these powerful players. To reduce these risks, a dual strategy can be implemented. On one side, this involves using more compact, open-source LLMs, which may, however, result in lower-quality outputs. On the other side, this entails using proprietary LLMs as software-as-a-service to end-users, which incurs financial expenses. For this research, we employed a mix of both strategies, highlighting their respective benefits and drawbacks.
    \item LLMs have no built-in understanding of the world, as they cannot directly experience or “simulate” our environment. This is unlike humans, who constantly perceive information, be it sensory, visual, or auditory. In contrast, LLM operation depends on the data with which they are trained. Consequently, if the data is not meticulously selected by humans, the model's predictions may be unreliable or even generate false information, a phenomenon known as “hallucinations” in AI discourse.\footnote{The term “hallucination” in the context of LLMs was derived from the concept of “AI hallucination,” which refers to incorrect responses generated by AI systems. In fact, this term was adopted due to the tendency of humans to anthropomorphize technology, attributing human qualities to it. As Floridi and Nobre have recently shown~\cite{Floridi2024-jv}, such a tendency is common with disruptive technologies. It has also been shown that, as we become more familiar with these technologies, the tendency to anthropomorphize should decrease over time.} We mitigated the problem of hallucinations with a special focus on prompt design. It has been observed that ambiguous prompts can lead to inconsistent results, with the same prompt potentially yielding different responses each time it is executed. As LLMs have evolved and increased in size, it has become evident that altering the prompt strategy can significantly improve the quality of the responses. This discovery has opened up new possibilities, enabling LLMs to tackle tasks that previously yielded negative or untested results. Carefully designing and adjusting prompts can improve the reliability and performance of LLMs~\cite{qiao2023reasoning}. This article focuses on creating prompts that produce desired outcomes.
\end{enumerate}

\section{Problem Definition}\label{sec:problem}
In this section, we provide the definition of the optimization problem we consider as an example in this paper, while in Section~\ref{sec:integration}, we present our hybrid optimization approach. The considered optimization problem is from the realm of social networks. In fact, social network problems often serve as an interesting experimental laboratory for testing optimization techniques since they can be modeled as combinatorial problems based on directed graphs that become highly complex as the instance size grows.

Specifically, we consider Multi-Hop Influence Maximization, a social network problem proven NP-hard by Ni et al. \cite{Ni} and Basuchowdhuri et al.~\cite{Basuchowdhuri}. In the literature, this problem has been studied using both metaheuristics and a state-of-the-art combination of a BRKGA with deep learning (DL)~\cite{9909110}, which gives us a point of comparison.

\Figure[h](topskip=0pt, botskip=0pt, midskip=0pt)[width=0.9\linewidth]{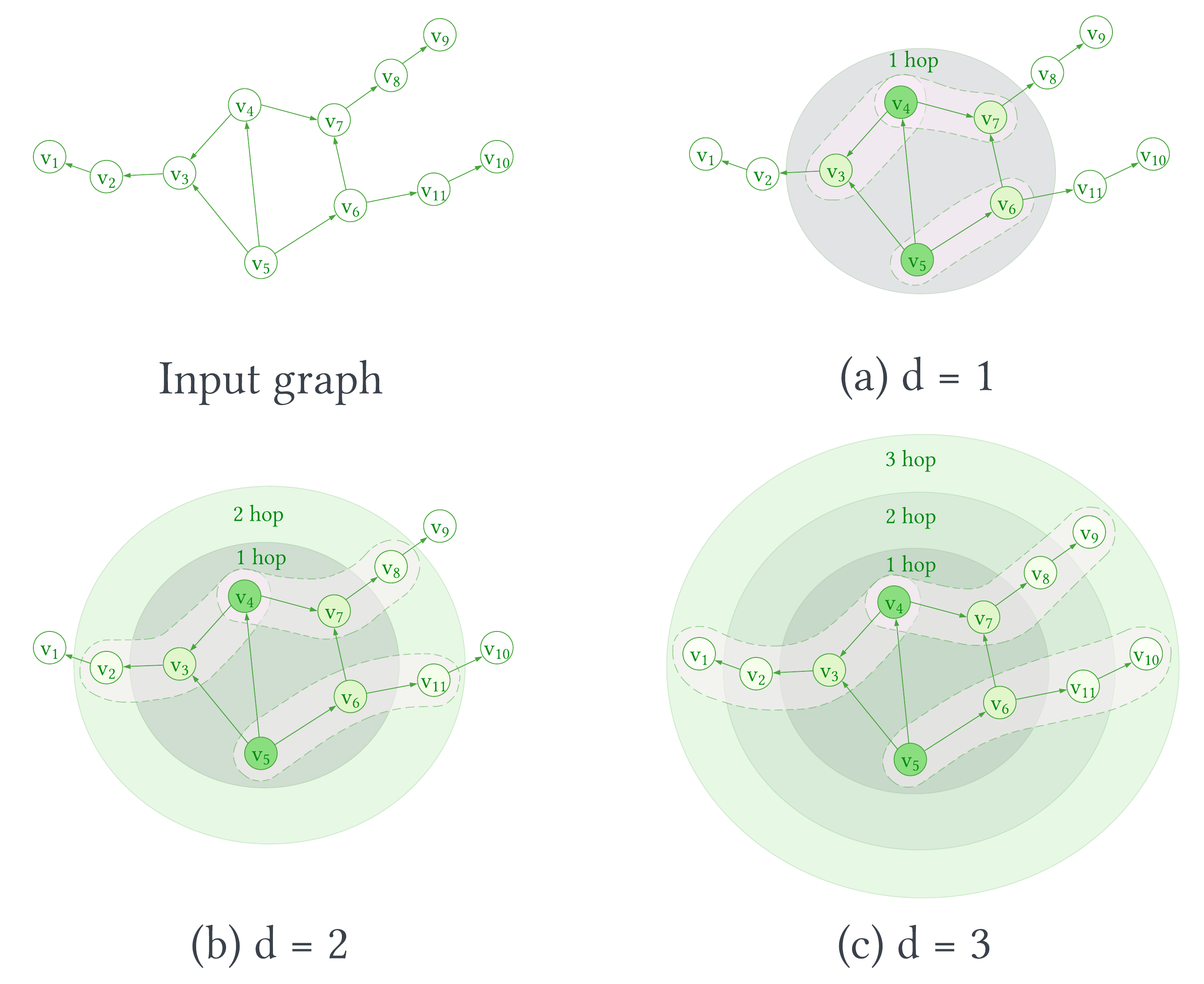}
{\textbf{Multi-hop influence process}. The given directed graph consists of 11 nodes and 12 arcs, and the task is to solve the $k$-$d$DSP with $k=2$. The example solution $U$ consists of two nodes: $v_4$ and $v_5$ (colored in green). The bottom row illustrates the concept of $d$-hop coverage: when $d=1$, nodes ${v_3, v_7, v_6}$ are 1-hop covered by $U$; when $d=2$, nodes ${v_2, v_3, v_7, v_8, v_6, v_{11}}$ are 2-hop covered by $U$; and when $d=3$, all remaining nodes in the graph are 3-hop covered by $U$.\label{fig:kdDSP}}

\subsection{Multi-Hop Influence Maximization in Social Networks}

Many optimization problems in social networks can be formalized by modeling the social network as a directed graph $G=(V,A)$, where $V$ represents the set of nodes and $A$ represents the set of directed arcs. This is also the case of the specific multi-hop influence maximization problem addressed in this paper, referred to as the $k$-$d$-Dominating Set Problem ($k$-$d$DSP).

The most crucial concept in this context is the \emph{influence} $I_d(u) \subseteq V$ of a node $u \in V$, which is determined by two factors:
\begin{enumerate}
\item Parameter $d \geq 1$, which is an input to the problem and represents the maximum distance of influence.
\item A distance measure $dist(u,v)$ between nodes $u$ and $v$. In this paper, $dist(u,v)$ is defined as the length (in terms of the number of arcs) of the shortest directed path from $u$ to $v$ in $G$.
\end{enumerate}
Based on these factors, we can define the influence of a node $u$ as follows:
\begin{equation}
I_d(u) := \{v \in V \: | \: dist(u, v) \leq d \}   
\end{equation}
In other words, $I_d(u)$ represents the set of all nodes in $G$ that can be reached from $u$ via a directed path with at most $d$ arcs. We say that $u$ influences (or covers) all nodes in $I_d(u)$. This definition can be naturally extended to sets of nodes as follows:
\begin{equation}\label{eq:influence}
I_d(U) := \bigcup\limits_{u \in U} I_{d}(u) \quad \forall U \subseteq V
\end{equation}
That is, $I_d(U)$ represents the set of all nodes in $G$ that are influenced by at least one node from the set $U$.

Valid solutions to the $k$-$d$DSP are all sets $U \subseteq V$ such that $|U| \leq k$, meaning that any valid solution can contain at most $k$ nodes. The objective of the $k$-$d$DSP is to find a valid solution $U^* \subseteq V$ such that $|I_d(U^*)| \geq |I_d(U)|$ for all valid solutions $U$ to the problem. In other words, the objective function value of a valid solution $U$ is $|I_d(U)|$. Formally, the $k$-$d$DSP can be stated as follows:
\begin{equation}
\begin{aligned}
\max_{U \subseteq V} \quad & |I_d(U)| \\
\textrm{s.t.} \quad & |U| \leq k \\
\end{aligned}
\end{equation}

For an intuitive explanation of $k$-$d$DSP, see the toy example in Figure~\ref{fig:kdDSP}.

\section{Integration of LLM Output into a Metaheuristic}\label{sec:integration}

\Figure[t!](topskip=0pt, botskip=0pt, midskip=0pt)[width=1\linewidth]{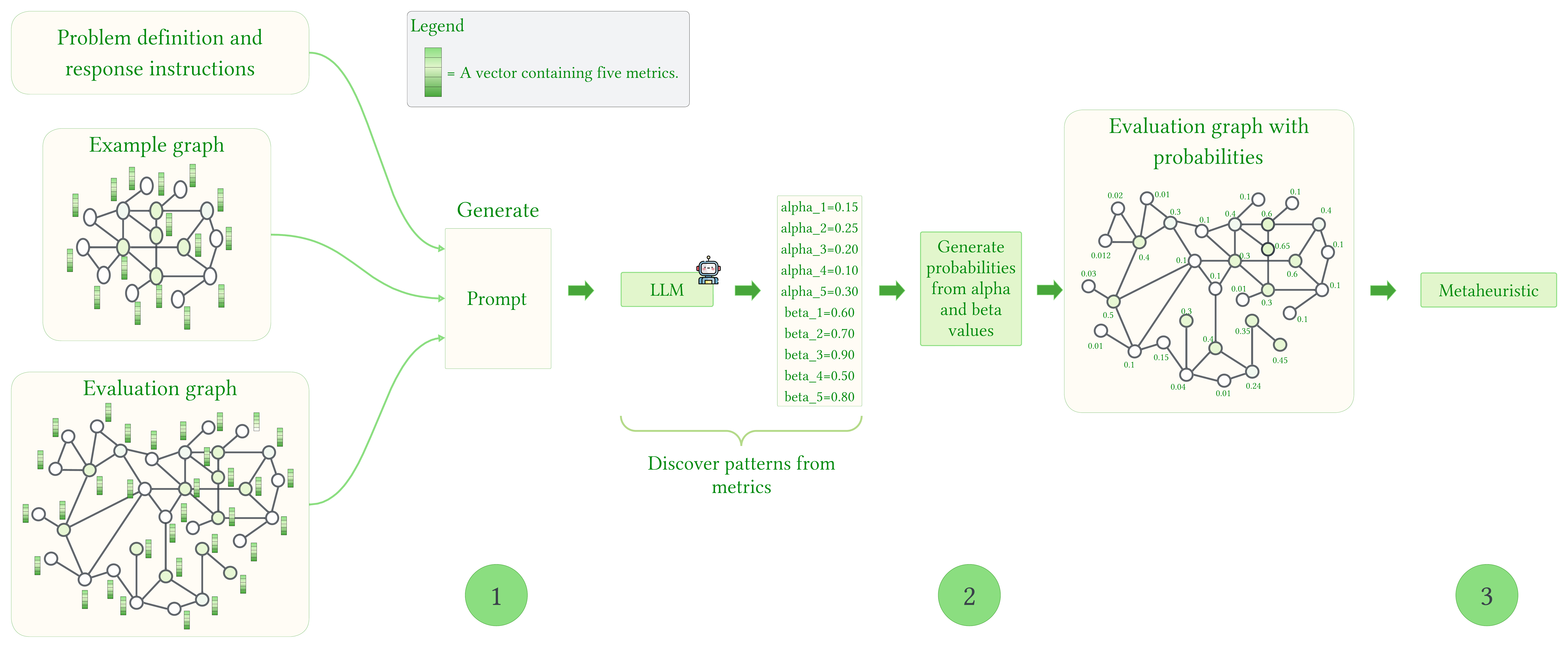}
{An overview of our approach to integrating MHs and LLMs: We employ LLMs to analyze problem instances and uncover hidden patterns. The patterns are then converted into useful information that guides the MH in its search for high-quality solutions.\label{fig:integration}}

Figure~\ref{fig:integration} depicts the framework of our proposed \texttt{MH+LLM} integration, comprising three automatic sequential steps:

\begin{enumerate}
    \item \textbf{Prompt generation and execution by an LLM.} We begin by phrasing the $k$-$d$DSP in natural text and creating a small random graph with a high-quality solution. Next, we calculate five key metrics for each node of the graph, which enables the LLM to determine the most relevant metrics for this problem. We then compute the same metrics for a second (larger) graph in which we want to solve the $k$-$d$DSP problem. This graph is henceforth called the \emph{evaluation graph}. Using the generated data, we design a prompt and ask the LLM to provide parameters for calculating the importance of each node in the evaluation graph. In essence, we leverage the LLM as a \textit{pattern recognition engine} to identify correlations between node metrics and node importance in the context of the $k$-$d$DSP.
    \item \textbf{Calculate probabilities for each node of the evaluation graph.} As explained in detail below, the LLM provides values for ten parameters that can be used to compute the probability of each node of the evaluation graph to form part of an optimal $k$-$d$DSP solution. We expect this information to offer excellent guidance to a MH.
    \item \textbf{Utilizing the probabilities (guidance) within a MH.} Since the MH we use in this work is a BRKGA, we incorporate the probabilities calculated based on the LLM output into the decoder that translates random keys into valid solutions to the tackled optimization problem.
\end{enumerate}

In what follows, these three steps are detailed in corresponding subsections.

\subsection{Prompt Engineering}\label{sec:prompt-engineering}

A prompt is an instruction provided as input to an LLM. The design of a prompt can greatly influence the quality of the LLMs' responses~\cite{wei2022emergent, wan2024efficient}. Also, the specific response can vary depending on the LLM used. From the available prompt design techniques, we have opted for one-shot learning~\cite{brown2020language}, also sometimes called few(1)-shot learning. According to Chen's findings~\cite{chen2023large}, tabular data reasoning can achieve good results with a single example, eliminating the need for additional examples or fine-tuning. This method involves furnishing the LLM with an example to facilitate pattern recognition in the response instructions that are to be requested.

\subsubsection{Definition}

The prompt we have designed consists of four tags, defined as follows:
\begin{equation}\label{eq:prompt}
P := \textsc{prompt}(\text{\textsf{Tag1}}, \text{\textsf{Tag2}}, \text{\textsf{Tag3}}, \text{\textsf{Tag4}})
\end{equation}
where
\begin{itemize}
    \item \textsf{Tag1} is the \textsf{[PROBLEM]} tag,
    \item \textsf{Tag2} is the \textsf{[EXAMPLE GRAPH]} tag,
    \item \textsf{Tag3} is the \textsf{[EVALUATION GRAPH]} tag, and
    \item \textsf{Tag4} is the \textsf{[RULES ANSWERING]} tag.
\end{itemize}
Hereby, the \textsf{[PROBLEM]} tag contains the description of the $k$-$d$DSP. Moreover, the example graph information is provided in the \textsf{[EXAMPLE GRAPH]} tag. Hereby, the example graph consists of 100 nodes, each characterized by the values of five metrics: \textsf{in-degree}, \textsf{out-degree}, \textsf{closeness}, \textsf{betweenness}, and \textsf{pagerank}.\footnote{
We selected these five metrics based on our understanding of the $k$-$d$DSP problem. However, in future work, leveraging the extensive knowledge LLMs possess from academic sources, we plan to use them to guide the selection of metrics for each specific problem. This could represent the extension of our method.} In particular, these values are henceforth denoted by
\begin{align}
 &m_{i,1}^{ex}, m_{i,2}^{ex}, m_{i,3}^{ex}, m_{i,4}^{ex}, m_{i,5}^{ex} \nonumber \\
 &\text{for all } v_i \text{ of the example graph.}    
\end{align}
Hereby, $m_{i,1}^{ex}$ is the value corresponding to metric \textsf{in-degree}, $m_{i,2}^{ex}$ corresponds to \textsf{out-degree}, etc. Note also that the metric values are normalized to the range $[0,1]$. Furthermore, the high-quality $k$-$d$DSP solution of the example graph is computed using the pure BRKGA algorithm, which we adopted from our earlier work~\cite{9909110}. The solution is encoded as a vector of 32 nodes, separated by commas, corresponding to the $k$-$d$DSP parameter $k=32$. \\

Next, the \textsf{[EVALUATION GRAPH]} tag contains the evaluation graph for which the $k$-$d$DSP must be solved. Each node of this graph is described by the values of the same five metrics described above. These evaluation graph values are henceforth denoted by 
\begin{align}
    & m_{i,1}^{eval}, m_{i,2}^{eval}, m_{i,3}^{eval}, 
     m_{i,4}^{eval}, m_{i,5}^{eval} \nonumber \\
    & \text{for all } v_i \text{ of the evaluation graph.}
\end{align}
Finally, the \textsf{[RULES ANSWERING]} tag specifies the details of the request to the LLM, which will be elaborated on in Section~\ref{subsec:prompt-structure}. \\

After the prompt $P$ is formulated, it is utilized by invoking the \textsc{execute} function, which takes three parameters: the prompt $P$, the selected $LLM$, and $\Theta$, representing a set of values for the configuration parameters of the $LLM$. This results in the corresponding LLM output:

\begin{equation}\label{eq:exec}
Output := \textsc{execute}(P, LLM,  \Theta)
\end{equation}

Specifically, $\Theta$ contains values for exactly two hyperparameters, regardless of the utilized LLM. The first, known as \textsf{temperature}, is a value between 0 and 1 that measures the model's response uncertainty, with lower values indicating a more deterministic output. While “more deterministic” does not imply that the LLM will generate identical responses to the same prompt every time, it does enhance the stability of the outputs, making it easier to replicate experiments~\cite{atil2024llmstabilitydetailedanalysis}. Therefore, we set the \textsf{temperature} to 0.\footnote{For text generation and paraphrasing tasks, it's recommended to increase the \textsf{temperature}. This adjustment promotes creativity, which is a desirable characteristic in these contexts. These applications differ from using the LLM as a pattern recognition tool, where consistent and stable outputs are preferable.} The second hyperparameter is the \textsf{maximum number of output tokens}, which we have set to a moderate \num{1000} tokens. This choice is based on the prompt design, which consistently yields relevant outputs regardless of the evaluation graph's size, ensuring that the quality of the results is not compromised by a smaller token limit.

\subsubsection{Prompt Structure}\label{subsec:prompt-structure}
Effective prompts are generally those with few language ambiguities. To achieve this, the four unique opening and closing tags mentioned in the previous section provide structure and coherence. We will now clarify the syntactic structure of each of these tags. A complete example of a prompt, along with each tag, can be found in Figure~\ref{fig:prompt}.

\Figure[p](topskip=0pt, botskip=0pt, midskip=0pt)[width=1\linewidth]{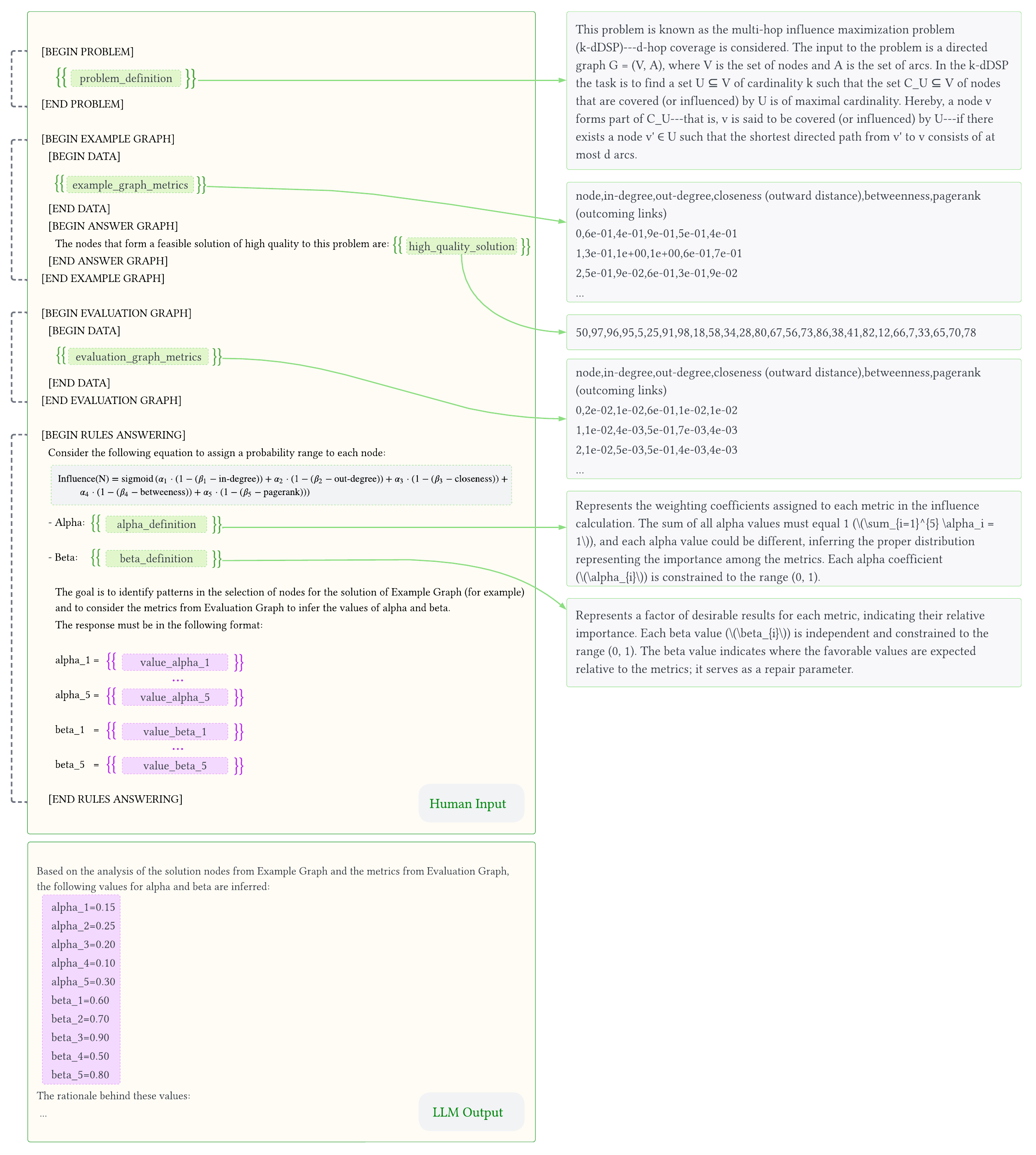}
{An example of a prompt and the corresponding LLM response. The prompt includes the problem definition, a graph example with node metrics and a high-quality solution, an evaluation graph, and instructions for the LLM for producing the output. Based on the patterns identified in the evaluation graph, the LLM provides the importance of each metric, represented by the set of alpha and beta values.\label{fig:prompt}}

\begin{enumerate}
    \item \textbf{Problem description.} The prompt starts by providing a concise definition of the $k$-$d$DSP utilizing LaTeX notation within the \textsf{[PROBLEM]} tag; see the top right of Figure~\ref{fig:prompt}.
    
    \item \textbf{Example Graph.} The \textsf{[EXAMPLE GRAPH]} tag, as the name suggests, provides information about the example graph. Nestled within this tag are two additional tags: \textsf{[DATA]}, encompassing the metric values of each node of the example graph, and \textsf{[ANSWER]}, which provides a high-quality solution for the given graph.

    \begin{itemize}
    \item \textsf{[DATA]} \textbf{tag}: A (directed) random graph with 100 nodes produced with the Erdös–Rényi model~\cite{paul1959random} was chosen as an example graph. The edge probability of the graph was $0.05$. Subsequently, five before-mentioned metrics were calculated for each of the 100 nodes and incorporated into the prompt in a tabular data format, with rows and columns separated by commas. Each row corresponds to a node ID, while the columns represent the respective metric values for that particular node. The metric values are presented in scientific numerical notation to minimize token usage. The rationale behind this decision is discussed in the context of the empirical results; see Section~\ref{sec:empirical-evaluation}.

    \item \textsf{[ANSWER]} \textbf{tag}: The solution to the example graph is computed using the BRKGA algorithm from~\cite{9909110}. However, note that this solution (which is not necessarily optimal) could potentially have been achieved through alternative means, such as employing a different metaheuristic or solving the problem via an exact method. The rationale behind including a high-quality solution is our expectation that---given the nodes belonging to a presumably high-quality solution---the LLM will be able to discern which metrics are more crucial than others and how the metric values of selected nodes interrelate.
    \end{itemize}
    
    \item \textbf{Evaluation Graph.} The \textsf{[EVALUATION GRAPH]} tag, much like the \textsf{[EXAMPLE GRAPH]} tag, utilizes a nested \textsf{[DATA]} tag to store the values of the five metrics for every node. However, we obviously do not provide any solution for the evaluation graph. This is because the objective is to request information from the LLM on the probability of nodes from the evaluation graph to pertain to an optimal solution.
    
    \item \textbf{Rules Answering.} The \textsf{[RULES ANSWERING]} tag is crucial as it ties together all the information provided in the previous tags. In this part of the prompt, an equation is presented to the LLM to calculate the probability of each node of the evaluation graph to form part of an optimal solution. The equation requires 10 parameters: 5 alpha parameters and 5 beta parameters, which will be explained in more detail in Section~\ref{subsec:alpha-beta}. These parameters serve to assign weights to the metrics and correct potential errors. The LLM infers the values of these parameters by analyzing the metrics in the \textsf{[EVALUATION GRAPH]} tag and using the metrics and the solution from the \textsf{[EXAMPLE GRAPH]} tag as a guide.
\end{enumerate}

For optimal prompt construction, we recommend using our \texttt{OptiPattern} tool, which generates the prompt automatically.

\subsection{LLM Output}\label{subsec:alpha-beta}

As described before, a prompt $P$ provides the values of the following five metrics for each node of the example graph and the evaluation graph: \textsf{in-degree}, \textsf{out-degree}, \textsf{closeness}, \textsf{betweenness}, and \textsf{pagerank}. It is assumed that the most important metric for addressing the $k$-$d$DSP is the \textsf{out-degree}, that is, the number of neighbors that can be reached from a node via directed arcs. A node with a higher out-degree is generally more likely to form part of high-quality solutions. However, we assume that there are additional metrics (among the other four metrics) that might contribute valuable information. Consequently, we anticipate that the LLM will be able to identify this. To identify patterns in the values provided by the metrics, the LLM is requested (by means of the \textsf{[RULES ANSWERING]} tag) to return values for two sets of five parameters (one for each metric, in the order as given above), resulting in ten values. More specifically, upon executing a prompt $P$, the chosen LLM produces a set $Output$ (see Eq.~(\ref{eq:exec})) which is as follows:

\begin{equation}
Output = \{\alpha_1, \ldots, \alpha_5, \beta_1, \ldots, \beta_5\}
\end{equation}

The first five of these values are henceforth called alpha values, while the last five values are named beta values. The heart concept of the proposed prompt is centered on the meaning of these values and how they are utilized. 

\begin{itemize}
    \item \textbf{alpha values}: These are weights that indicate the influence of each metric. The total sum of all alpha values should be equal to one ($\sum_{i=1}^{5} \alpha_i = 1$), and each alpha value can be unique. In other words, the alpha value $0 < \alpha_i < 1$ reflects the relative significance of the $i$-th metric (in the order as mentioned above).
    \item \textbf{beta values}: These five values are adjustment (or correction) parameters. Unlike the alpha values, beta values $0 < \beta_i < 1$ are independent of each other. Moreover, beta values do not represent relative weights among the metrics. They rather indicate the best possible value of a node regarding a metric. This allows the LLM to identify where the best values are found with respect to their range $[0, 1]$.
    
    
\end{itemize}

Based on these values from the LLM output, the probability for a node $v_j$ of the evaluation graph is determined using the following formula: 
\begin{equation}\label{eq:influence-llm}
\begin{aligned}
\mathbf{p}^{\mathrm{LLM}}(v_j) := \sigma\Bigg(& \sum_{i=1}^{5}{\alpha_i \cdot (1 - (\beta_i - m_{j,i}^{eval}))}   \Bigg)
\end{aligned}
\end{equation}

Note that this formula introduces non-linearity into the node probabilities by applying the sigmoid function $\sigma$, which enables a more nuanced representation of the probability space.\footnote{The sigmoid function has been used for many purposes in neural networks. But also in metaheuristics, for example, for significantly accelerating the convergence of a genetic algorithm~\cite{4344687}.} 

As shown in Figure~\ref{fig:prompt}, our proposed prompt thoroughly explains the alpha and beta values to the LLM, along with Eq.~\ref{eq:influence-llm}. By giving the LLM a clear understanding of the context surrounding the alpha and beta values, we simply ask the LLM to provide the corresponding values for the evaluation graph.

\subsection{Using LLM Output to Guide a Metaheuristic}

In this section, we first describe the metaheuristic considered to test the quality of the LLM output. Subsequently, the way of incorporating the probability values into the metaheuristic is outlined.

\subsubsection{The Considered Metaheuristic: A BRKGA}
\label{sec:BRKGA}

The chosen metaheuristic is a so-called Biased Random Key Genetic Algorithm (BRKGA) designed for solving the $k$-$d$DSP in~\cite{9909110}.\footnote{BRKGA's are well-known GA variants, mostly for solving combinatorial optimization problems.} In fact, two algorithm variants were proposed in~\cite{9909110}: (1) a pure BRKGA variant and (2) an algorithm variant that makes use of a hand-designed deep learning framework for biasing the BRKGA. This will allow us to compare our proposal properly to existing algorithm versions. 

In general, a BRKGA is problem-independent because it works with populations of individuals that are vectors of real numbers (random keys). The problem-dependent part of each BRKGA deals with how individuals are translated into solutions to the tackled problem. The problem-independent pseudocode of BRKGA is provided in Algorithm~\ref{alg:brkga}. 

The algorithm begins by calling \textsc{GenerateInitialPopulation}($p_{size}$, $seed$) to create a population $P$ of $p_{size}$ individuals. If $seed = 0$, all individuals are randomly generated, with each $\pi \in P$ being a vector of length $|V|$ (where $V$ is the set of nodes from the input graph). The value at position $i$ of $\pi$, $\pi(i)$, is randomly chosen from $[0, 1]$ for all $i=1,\ldots,|V|$. If $seed = 1$, $p_{size}-1$ individuals are randomly generated, and the last individual is obtained by setting $\pi(i) := 0.5$ for all $i=1,\ldots,|V|$. The initial population's individuals are then evaluated by transforming each individual $\pi \in P$ into a valid solution $U_{\pi} \subset V$ to the $k$-$d$DSP, with the value $f(\pi)$ defined as $f(\pi) := |U_{\pi}|$. The transformation process is discussed later.

At each iteration, the algorithm performs the following operations:
\begin{enumerate}
    \item The best max$\{\lfloor p_e \cdot p_{size} \rfloor, 1\}$ individuals are copied from $P$ to $P_e$ using \textsc{EliteSolutions}($P, p_e$).
    \item A set of max$\{\lfloor p_m \cdot p_{size}\rfloor, 1\}$ mutants are generated by function \textsc{Mutants}$(P, p_m)$ and stored in $P_m$. These mutants are random individuals generated the same way as those from the initial population.
    \item A set of $p_{size} - |P_e| - |P_m|$ individuals are generated by crossover using \textsc{Crossover}($P,p_e,prob_{elite}$) and stored in $P_c$. The crossover, which involves combining two solutions, serves as the mechanism that enhances the search process, concentrating on transferring the superior traits of parents to their offspring.
\end{enumerate}

\algrenewcommand\algorithmicfunction{}

\begin{algorithm}[!t]
\caption{The pseudocode of BRKGA}\label{alg:brkga}
\begin{algorithmic}[1] 
\Require \text{a directed graph $G = (V, E)$}
\Ensure \text{values for parameters~$p_{size}$, $p_e$, $p_m$, $prob_{elite}$, $seed$}

\State $P \gets {\small \Call{GenerateInitialPopulation}{p_{size}, seed}}$

\State {\small\Call{Evaluate}{$P$}} \Comment{problem-dependent part (greedy)}

\While{\text{computation time limit not reached}}

\State $P_e \gets {\small\Call{EliteSolutions}{P, p_e}}$
\State $P_m \gets {\small\Call{Mutants}{P, p_m}}$
\State $P_c \gets {\small\Call{Crossover}{P, p_e, prob_{elite}}}$

\State {\small\Call{Evaluate}{$P_m \cup P_c$}} \Comment{problem-dependent part (greedy)}
\State $P \gets P_e \cup P_m \cup P_c$

\EndWhile

\State \Return \text{Best solution in $P$}
\end{algorithmic}
\end{algorithm}

The evaluation of an individual (see lines~2 and~7 of Algorithm~\ref{alg:brkga}) is the crucial problem-dependent aspect of the BRKGA algorithm from~\cite{9909110}. This evaluation function---often termed the \textit{decoder}---utilizes a straightforward greedy heuristic. The heuristic is based on the notion that nodes with a higher out-degree---that is, more neighbors---are likely to yield a higher influence. 

For a node $v_j \in V$, the set of neighbors, $N(v_j)$, comprises nodes reachable via a directed arc from $v_j$: $N(v_j) = \{v_i \in V \mid (v_j, v_i) \in A\}$. The greedy value $\phi(v_j)$ for each $v_j \in V$ is calculated as:

\begin{equation}\label{eq:greedy}
    \phi(v_j) := |N(v_j)| \cdot \pi(j)
\end{equation}

In other words, in this equation, the greedy value of a node $v_j$ is determined by multiplying its out-degree with the corresponding numerical value from the individual being translated into a solution. The final solution, $U_{\pi}$, is obtained by selecting the $k$ nodes with the highest greedy values.

The following will modify the greedy function $\phi$ to create our hybrid algorithm.

\subsubsection{Hybrid algorithm}\label{sec:hybrid}

The proposed hybrid algorithm---referred to as \texttt{BRKGA+LLM}---begins with two offline steps. Given an evaluation graph $G=(V,A)$, a prompt is generated as outlined in the previous section and sent to an LLM. Based on the alpha and beta values obtained from the LLM's response, the probability $\mathbf{p}^\mathrm{LLM}(v_j)$ for each node $v_j \in V$ is determined using Eq.~(\ref{eq:influence-llm}). Next, the original greedy function $\phi()$ from Eq.~(\ref{eq:greedy}) is substituted with a modified version that integrates the node probabilities derived from Eq.~(\ref{eq:influence-llm}):

\begin{equation}\label{eq:hybrid}
    \phi_{\textsc{Influence}_{LLM}}(v_j) := |N(v_j)| \cdot \pi(j) \cdot \mathbf{p}^\mathrm{LLM}(v_j) \quad \forall v_j \in V
\end{equation}

We hypothesize that with suitable predictions from the LLM, the algorithm can be guided/biased to explore more promising areas of the search space. These areas are believed to contain high-quality solutions that the BRKGA would have been unable to discover on their own without the guidance of these predictions. In other words, using an LLM to discover patterns in metric values (see Section~\ref{sec:prompt-engineering}), rather than relying solely on the out-degree, might enhance the algorithm's performance. 

\section{Empirical Evaluation}\label{sec:empirical-evaluation}

This section presents empirical evidence demonstrating the benefits of integrating MHs and LLMs. The following algorithm variants are considered for the comparison:
\begin{itemize}
    \item \texttt{BRKGA}: the pure BRKGA variant already published in~\cite{9909110} and described above in Section~\ref{sec:BRKGA}.
    \item \texttt{BRKGA+FC}: the BRKGA hybridized with a hand-designed GNN called \texttt{FastCover} (FC) that was used to derive the probability values (last term of Eq.~(\ref{eq:hybrid})) in~\cite{9909110}.
    \item \texttt{BRKGA+LLM}: the BRKGA enhanced with LLM output as described in the previous section.
\end{itemize}
Note that both \texttt{BRKGA} and \texttt{BRKGA+FC} underwent a general parameter tuning in~\cite{9909110} depending on the value of $k$. The well-known \texttt{irace} tool~\cite{lopez2016irace} was used for this purpose. In this work, we adopt the corresponding parameter settings of \texttt{BRKGA} for \texttt{BRKGA+LLM}. In this way, we can be sure that any difference in their performance is caused by the guidance of the probabilities computed from the LLM outputs. In any case, a specific tuning of \texttt{BRKGA+LLM} could only further improve its results. 

Apart from comparing the three approaches mentioned above, we show results for different LLMs and provide evidence for the quality of LLM output. Additionally, we support our analysis with a visual examination, providing additional insight into why the hybrid \texttt{BRKGA+LLM} outperforms the other algorithm variants.

\begin{table*}[!t]\centering
\caption{Summary of the assessed LLMs, which have been used via the OpenRouter API. This is except for Claude-3-Opus, the first LLM considered. At that point, we had yet to become familiarized with OpenRouter.}\label{table:overview-models}
\resizebox{\linewidth}{!}{

\begin{tabular}{lcccccc}\toprule
Model & Chatbot & Version & License & Maximum & Test Environment (API) \\
      & Arena   &         &         & Context & \\
      & Ranking &         &         & Window  &
   \\ \midrule
OpenAI/GPT-4o &\#1 &may2024 &private &\num{128000} &OpenRouter \\
Anthropic/Claude-3-Opus &\#2 &march2024 &private &\num{200000} &Anthropic \\
Cohere/Command-R+ &\#10 &april2024 &CC-BY-NC-4.0  &\num{128000 } &OpenRouter \\
MistralAI/Mixtral-8x22b-Instruct-v0.1 &\#19 &april2024 &Apache 2.0 &\num{32768} &OpenRouter  \\

\bottomrule
\end{tabular}
}
\end{table*}

\subsection{Experimental Setup}

The BRKGA was implemented in \texttt{C++}, whereas the prompt construction process, which entails extracting metrics from graph instances, was conducted using Python 3.11. Regarding the choice of LLMs, we utilized two proprietary language models, GPT-4o and Claude-3-Opus, as well as two open-source models, Command-R+ and Mixtral-8x22b-Instruct-v0.1. We selected these models based on the Chatbot Arena---a platform developed by LMSYS members and UC Berkeley SkyLab researchers---which provides an Arena 
Leaderboard,\footnote{\url{https://lmarena.ai}.} a community-driven ranking system for LLMs~\cite{zheng2023judging}.\footnote{Please be aware that our experiments took place between February and May 2024, and the LLMs ranking classification in Chatbot Arena may have changed by the time of reading.} Table~\ref{table:overview-models} presents a comprehensive overview of the models, including their ranking in the Chatbot Arena Leaderboard (as of May 2024), corresponding version numbers, licenses, maximum context windows, and crucially, the test environment employed for each model. 

\subsubsection{Execution Environment}

We utilized the OpenRouter API\footnote{\url{https://openrouter.ai}.} to execute prompts in their corresponding LLMs, except for Claude-3-Opus, which we used through the Anthropic API (see Table~\ref{table:overview-models}). Finally, all experiments involving the three BRKGA variants were conducted on a high-performance computing cluster comprising machines powered by Intel Xeon CPU 5670 processors with 12 cores running at 2.933 GHz and a minimum of 32 GB of RAM.

\subsubsection{Dataset}\label{subsection:dataset-restriction}

Our evaluation is based on two categories of $k$-$d$DSP instances (evaluation graphs). The first consists of rather small, synthetic social network graphs with \num{500} and \num{1000} nodes, generated using three configuration methods developed by Nettleton~\cite{Nettleton2016}. The corresponding graph generator requires four real-valued parameters, whose values are reflected by the instance names.\footnote{The instance names are obtained by a concatenation of the utilized parameter values: examples are \textsf{0.4-0.15-0.15-0.3}, \textsf{0.3-0.0-0.3-0.4}, and \textsf{0.2-0.0-0.3-0.5}. The parameters labeled $a$, $b$, $c$, and $d$, respectively, define communities weights ($a$ and $d$) and link weights between communities ($b$ and $c$), $a+b+c+d\approx1$. These parameters influence the topology of the network, specifically the total number of connections and the density.} The second instance set comprises four real-world social network graph instances obtained from the well-established SNAP (Stanford Network Analysis Project) repository~\cite{snapnets}. Moreover, note that the $k$-$d$DSP can be solved in each graph for different values of $d$ and $k$. In this work we solved all evaluation graphs with $d \in \{1, 2, 3\}$ and $k \in \{32, 64, 128\}$. All datasets (synthetic and real), prompts, and results can be found in the \texttt{supplementary material/} folder in the repository: \url{https://github.com/camilochs/optipattern}. \\


\textbf{Restrictions.} The size of the graphs poses a constraint on the prompts we have designed for the LLMs, which is limited by two factors:

\begin{enumerate}
    \item The maximum context window of LLMs is still relatively small.\footnote{The maximum context window of an LLM sets the maximum amount of text it can process simultaneously when generating a response. This constraint determines the scope of contextual information the LLM can draw upon when answering a question or completing a task. The response quality will likely degrade if the input prompt exceeds this limit. Given that our prompt design requires each metric for each node to be equally important, the LLM needs to consider as much context as possible to deliver reasonable and trustworthy results.} For instance, the largest evaluation graph we use, \textsf{soc-wiki-elec}, results in an input prompt size of \num{181719} tokens, which is close to the \num{200000} tokens limit of Claude-3-Opus~\cite{Anthropic2024-ee}, the LLM which offers the currently largest context window. 
    \item The cost of processing larger instances is prohibitively high. For example, executing the prompt regarding the \textsf{soc-wiki-elec} evaluation graph on Claude-3-Opus exceeds €2.5.
\end{enumerate}

Table~\ref{table:models-costs} provides a detailed breakdown of the constraints for the largest evaluation graphs considered in this work. Although these limitations currently restrict us to testing with smaller instances, we anticipate that this constraint will soon be alleviated as the maximum context window increases and processing costs decrease (see Section~\ref{sec:discussion} for more information on this).

\begin{table}[!t]\centering
\caption{Number of input/output tokens and the associated cost of processing the input prompts concerning Claude-3-Opus. The costs correspond to March 2024.}\label{table:models-costs}
\scriptsize
\begin{tabular}{lcccc}\toprule
Instance &Input (tokens) &Output (tokens) &Cost (USD/EUR) \\\midrule
\textsf{soc-wiki-elec} &\num{181719} &\num{463} &2,78/2,58 \\
\textsf{soc-advogato} &\num{160812} &\num{410} &2,46/2,28 \\
\textsf{sign-bitcoinotc} &\num{66406} &\num{303} &1,02/0,95 \\
\textsf{soc-hamsterster} &\num{17097} &\num{438} &0,29/0,27 \\
\bottomrule
\end{tabular}
\end{table}

\subsection{Analysis of LLM Output}\label{emp:analysis}

\Figure[t!](topskip=0pt, botskip=0pt, midskip=0pt)[width=1\linewidth]{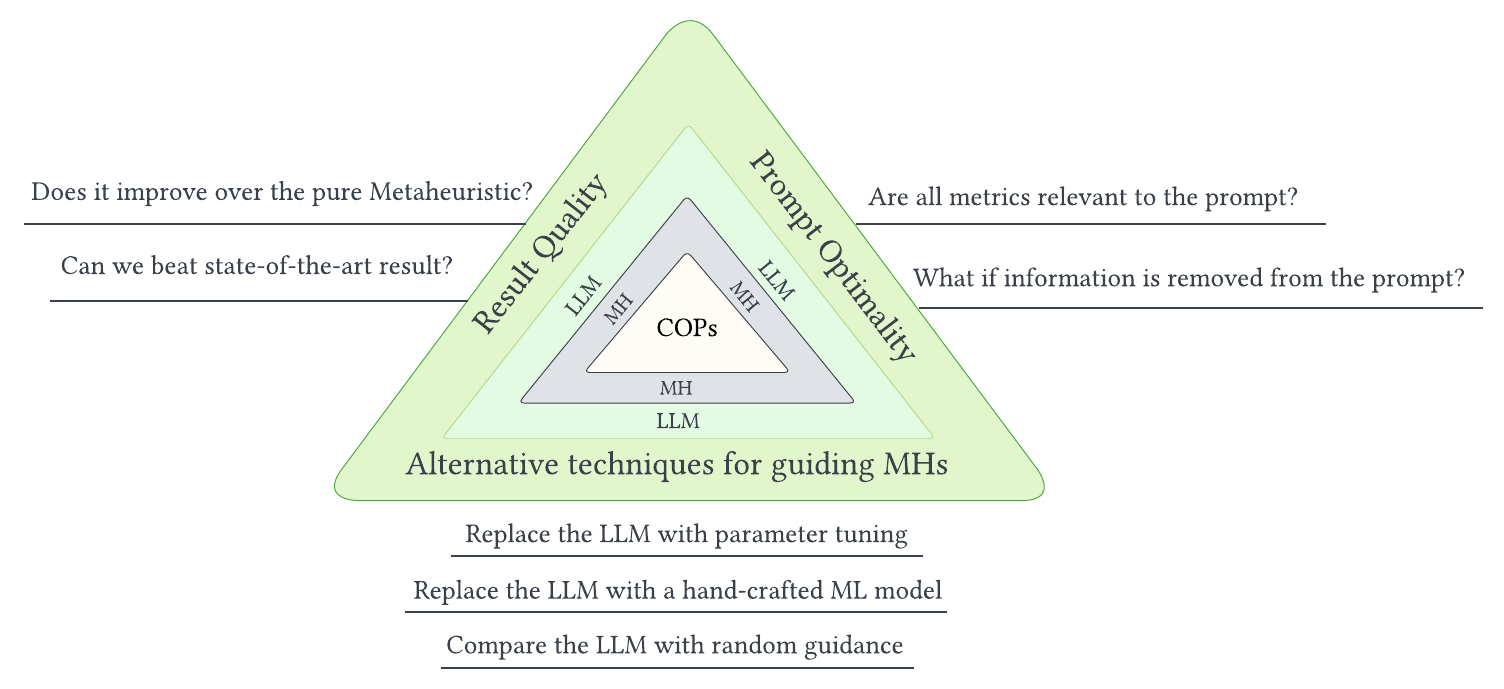}
{A comprehensive evaluation framework was used to assess the usefulness of integrating MHs with LLMs for solving combinatorial optimization problems (COPs) across the three dimensions shown in the graphic.\label{fig:framework}}

This section aims to validate the outputs of the LLMs and their usefulness for utilizing them as guidance within the BRKGA algorithm. The aim is to demonstrate that they are not arbitrary or devoid of significance. We have conducted three sets of experiments to achieve this, aiming at different aspects. Figure~\ref{fig:framework} shows the custom-designed, three-dimensional experimental framework developed specifically for this evaluation.\footnote{Future research could expand the framework's dimensions to better justify LLMs' response quality and integration with MHs and to address unexplored aspects.} 


\begin{table*}[!t]
\caption{Solution qualities obtained when turning the probabilities computed based on the LLM's output directly into solutions. In addition, the same is done for the \textbf{out-degree} metric. Considered LLMs are GPT-4o, Claude-3-Opus, Command-R+, and Mixtral-8x22b-Instruct-v0.1. The six synthetic graphs are chosen as a test bed. Green cells highlight the best LLM results, while gray cells indicate the top \textbf{out-degree} metric results.} \label{table:select-llm}
\scalebox{1}{
\begin{tabular}{lccc|ccccc|ccccc|ccccc}\toprule
& & & &\multicolumn{5}{c}{\textbf{$k=32$}} &\multicolumn{5}{c}{\textbf{$k=64$}} &\multicolumn{5}{c}{\textbf{$k=128$}} \\\cmidrule{5-19}
\textbf{Instance} &$|V|$ &$|E|$ & $d$ &\rotatebox{90}{\textbf{GPT-4o}} &\rotatebox{90}{\textbf{Claude-3-Opus}} &\rotatebox{90}{\textbf{Command-R+}} &\rotatebox{90}{\textbf{Mixtral-8x22b}} &\rotatebox{90}{\textsf{out-degree}} &\rotatebox{90}{\textbf{GPT-4o}} &\rotatebox{90}{\textbf{Claude-3-Opus}} &\rotatebox{90}{\textbf{Command-R+}} &\rotatebox{90}{\textbf{Mixtral-8x22b}} &\rotatebox{90}{\textsf{out-degree}} &\rotatebox{90}{\textbf{GPT-4o}} &\rotatebox{90}{\textbf{Claude-3-Opus}} &\rotatebox{90}{\textbf{Command-R+}} &\rotatebox{90}{\textbf{Mixtral-8x22b}} &\rotatebox{90}{\textsf{out-degree}} \\\midrule 
\multirow{3}{*}{0.4-0.15-0.15-0.3} &\multirow{3}{*}{500} &\multirow{3}{*}{3000} &1 &240 &241 &240 &241 &\cellcolor[HTML]{efefef}\textbf{256} &338 &339 &332 &330 &\cellcolor[HTML]{efefef}\textbf{361} &425 &428 &430 &428 &\cellcolor[HTML]{efefef}\textbf{439} \\
& & &2 &458 &458 &\cellcolor[HTML]{e2f6cc}\textbf{461} &457 &\cellcolor[HTML]{efefef}\textbf{461} &481 &481 &483 &479 &\cellcolor[HTML]{efefef}\textbf{485} &\cellcolor[HTML]{e2f6cc}\textbf{494} &\cellcolor[HTML]{e2f6cc}\textbf{494} &\cellcolor[HTML]{e2f6cc}\textbf{494} &\cellcolor[HTML]{e2f6cc}\textbf{494} &492 \\
& & &3 &\cellcolor[HTML]{e2f6cc}\textbf{494} &\cellcolor[HTML]{e2f6cc}\textbf{494} &\cellcolor[HTML]{e2f6cc}\textbf{494} &\cellcolor[HTML]{e2f6cc}\textbf{494} &493 &494 &495 &\cellcolor[HTML]{e2f6cc}\textbf{496} &494 &495 &\cellcolor[HTML]{e2f6cc}\textbf{496} &\cellcolor[HTML]{e2f6cc}\textbf{496} &\cellcolor[HTML]{e2f6cc}\textbf{496} &\cellcolor[HTML]{e2f6cc}\textbf{496} &\cellcolor[HTML]{efefef}\textbf{496} \\ \cmidrule{5-19}
\multirow{3}{*}{0.3-0.0-0.3-0.4} &\multirow{3}{*}{500} &\multirow{3}{*}{3000} &1 &302 &316 &308 &288 &\cellcolor[HTML]{efefef}\textbf{323} &391 &\cellcolor[HTML]{e2f6cc}\textbf{393} &392 &375 &384 &435 &\cellcolor[HTML]{e2f6cc}\textbf{449} &439 &429 &448 \\
& & &2 &366 &377 &371 &358 &\cellcolor[HTML]{efefef}\textbf{379} &416 &\cellcolor[HTML]{e2f6cc}\textbf{421} &418 &404 &410 &442 &\cellcolor[HTML]{e2f6cc}\textbf{455} &447 &439 &453 \\
& & &3 &369 &380 &374 &360 &\cellcolor[HTML]{efefef}\textbf{381} &416 &\cellcolor[HTML]{e2f6cc}\textbf{421} &418 &404 &410 &442 &\cellcolor[HTML]{e2f6cc}\textbf{455} &447 &439 &453 \\ \cmidrule{5-19}
\multirow{3}{*}{0.2-0.0-0.3-0.5} &\multirow{3}{*}{500} &\multirow{3}{*}{3000} &1 &164 &\cellcolor[HTML]{e2f6cc}\textbf{168} &163 &163 &155 &236 &\cellcolor[HTML]{e2f6cc}\textbf{250} &244 &244 &225 &321 &\cellcolor[HTML]{e2f6cc}\textbf{332} &331 &331 &297 \\
& & &2 &198 &\cellcolor[HTML]{e2f6cc}\textbf{202} &201 &201 &179 &253 &\cellcolor[HTML]{e2f6cc}\textbf{269} &264 &264 &237 &328 &339 &\cellcolor[HTML]{e2f6cc}\textbf{340} &\cellcolor[HTML]{e2f6cc}\textbf{340} &301 \\
& & &3 &202 &\cellcolor[HTML]{e2f6cc}\textbf{206} &203 &203 &179 &254 &\cellcolor[HTML]{e2f6cc}\textbf{269} &264 &264 &237 &328 &339 &\cellcolor[HTML]{e2f6cc}\textbf{340} &\cellcolor[HTML]{e2f6cc}\textbf{340} &301 \\ \cmidrule{5-19}
\multirow{3}{*}{0.4-0.15-0.15-0.3} &\multirow{3}{*}{1000} &\multirow{3}{*}{8000} &1 &381 &379 &378 &n.a. &\cellcolor[HTML]{efefef}\textbf{392} &534 &551 &552 &n.a. &\cellcolor[HTML]{efefef}\textbf{560} &727 &732 &734 &n.a. &\cellcolor[HTML]{efefef}\textbf{748} \\
& & &2 &926 &925 &925 &n.a. &\cellcolor[HTML]{efefef}\textbf{931} &\cellcolor[HTML]{e2f6cc}\textbf{973} &971 &971 &n.a. &\cellcolor[HTML]{efefef}\textbf{973} &988 &988 &989 &n.a. &\cellcolor[HTML]{efefef}\textbf{990} \\
& & &3 &992 &992 &991 &n.a. &\cellcolor[HTML]{efefef}\textbf{993} &993 &993 &993 &n.a. &\cellcolor[HTML]{efefef}\textbf{994} &994 &\cellcolor[HTML]{e2f6cc}\textbf{995} &\cellcolor[HTML]{e2f6cc}\textbf{995} &n.a. &\cellcolor[HTML]{efefef}\textbf{995} \\ \cmidrule{5-19}
\multirow{3}{*}{0.3-0.0-0.3-0.4} &\multirow{3}{*}{1000} &\multirow{3}{*}{8000} &1 &523 &561 &509 &n.a. &\cellcolor[HTML]{efefef}\textbf{571} &701 &720 &692 &n.a. &\cellcolor[HTML]{efefef}\textbf{723} &823 &\cellcolor[HTML]{e2f6cc}\textbf{848} &819 &n.a. &842 \\
& & &2 &743 &\cellcolor[HTML]{e2f6cc}\textbf{762} &728 &n.a. &755 &804 &810 &799 &n.a. &\cellcolor[HTML]{efefef}\textbf{812} &857 &\cellcolor[HTML]{e2f6cc}\textbf{880} &855 &n.a. &874 \\
& & &3 &758 &773 &747 &n.a. &\cellcolor[HTML]{efefef}\textbf{775} &808 &815 &803 &n.a. &\cellcolor[HTML]{efefef}\textbf{819} &858 &\cellcolor[HTML]{e2f6cc}\textbf{880} &856 &n.a. &874 \\ \cmidrule{5-19}
\multirow{3}{*}{0.2-0.0-0.3-0.5} &\multirow{3}{*}{1000} &\multirow{3}{*}{8000} &1 &232 &\cellcolor[HTML]{e2f6cc}\textbf{237} &230 &n.a. &223 &343 &\cellcolor[HTML]{e2f6cc}\textbf{349} &341 &n.a. &316 &501 &\cellcolor[HTML]{e2f6cc}\textbf{512} &\cellcolor[HTML]{e2f6cc}\textbf{512} &n.a. &470 \\
& & &2 &308 &311 &\cellcolor[HTML]{e2f6cc}\textbf{322} &n.a. &269 &394 &399 &\cellcolor[HTML]{e2f6cc}\textbf{402} &n.a. &343 &526 &535 &\cellcolor[HTML]{e2f6cc}\textbf{537} &n.a. &482 \\
& & &3 &313 &320 &\cellcolor[HTML]{e2f6cc}\textbf{328} &n.a. &272 &396 &401 &\cellcolor[HTML]{e2f6cc}\textbf{405} &n.a. &345 &526 &535 &\cellcolor[HTML]{e2f6cc}\textbf{537} &n.a. &482 \\
\bottomrule
\end{tabular}}
\end{table*}

Before starting with the main experimental evaluation, we must choose the most suitable LLM for the considered task. For this purpose, we produced all prompts concerning the six synthetic graphs from the dataset (as described before) for all considered combinations of $d \in \{1, 2, 3\}$  and $k \in \{32, 64, 128\}$. These prompts were fed into the following LLMs: GPT-4o, Claude-3-Opus, Command-R+, and Mixtral-8x22b-Instruct-v0.1. The obtained probabilities were then directly used to produce solutions containing the $k$ nodes with the largest probability values. The same was done concerning metric \textsf{out-degree}. That is, solutions were produced that consist of the $k$ nodes with the highest \textsf{out-degree} values. The results shown in Table~\ref{table:select-llm} allow us to make the following observations. First, although no LLM always outperforms the other LLMs, Claude-3-Opus shows advantages over the other LLMs, especially for increasing values of $k$. When comparing the results obtained with \textsf{out-degree} (the most popular greedy heuristic for the $k$-$d$DSP) to the results obtained with Claude-3-Opus, we can observe that \textsf{out-degree} seems to work slightly better for $k=32$, while the opposite is the case for $k=128$. Based on these results, we use Claude-3-Opus for the remainder of our experiments.

\subsubsection{Dimension 1 of the Evaluation Framework: Result Quality}


In the first set of experiments, we decided to compare the pure \texttt{BRKGA} approach with \texttt{BRKGA+LLM} in the context of the six synthetic graphs (and for all combinations of $d \in \{1, 2, 3\}$  and $k \in \{32, 64, 128\}$). Both algorithms were applied 10 times to each case, with a computation time limit of 900 CPU seconds per run. The results, which are shown in Table~\ref{table:synthetic-brkga-llm}, clearly show that \texttt{BRKGA+LLM} outperforms the pure \texttt{BRKGA} algorithm most of the time. Considerable improvements can be observed in the context of the last graph; see instance \textsf{0.2-0.0-0.3-0.5} with \num{1000} nodes and \num{8000} arcs. Only in three cases, the result obtained by \texttt{BRKGA+LLM} is slightly inferior to the one of \texttt{BRKGA}. While these results are promising, it is important to recognize that they are based on relatively small instances. 

\begin{table*}[h]
\caption{Comparison of the pure \texttt{BRKGA} with \texttt{BRKGA+LLM} on the six synthetic social networks. For each network, the algorithms were applied for each combination of $d \in \{1, 2, 3\}$ and $k \in \{32, 64, 128\}$. Average results over 10 algorithm runs are shown. Green cells indicate the best quality metrics results---higher values are better in this maximization problem.} \label{table:synthetic-brkga-llm}
\resizebox{\linewidth}{!}{
\begin{tabular}{cccc|cc|cc|cc}\toprule
& & & &\multicolumn{2}{c}{\textbf{$k=32$}} &\multicolumn{2}{c}{\textbf{$k=64$}} &\multicolumn{2}{c}{\textbf{$k=128$}} \\\cmidrule{5-10}
\textbf{Instance} &\textbf{$|N|$} &\textbf{$|E|$} &\textbf{$d$} &\texttt{BRKGA} &\texttt{BRKGA+LLM} &\texttt{BRKGA} &\texttt{BRKGA+LLM} &\texttt{BRKGA} &\texttt{BRKGA+LLM} \\\midrule
\multirow{3}{*}{0.4-0.15-0.15-0.3} &\multirow{3}{*}{500} &\multirow{3}{*}{3000} &1 &309.6 &\cellcolor[HTML]{e2f6cc}\textbf{309.9} &433.9 &\cellcolor[HTML]{e2f6cc}\textbf{437.2} &499.0 &\cellcolor[HTML]{e2f6cc}\textbf{500.0} \\
& & &2 &496.0 &\cellcolor[HTML]{e2f6cc}\textbf{499.9} &497.4 &\cellcolor[HTML]{e2f6cc}\textbf{500.0} &\cellcolor[HTML]{e2f6cc}\textbf{500.0} &\cellcolor[HTML]{e2f6cc}\textbf{500.0} \\
& & &3 &496.0 &\cellcolor[HTML]{e2f6cc}\textbf{500.0} &498.0 &\cellcolor[HTML]{e2f6cc}\textbf{500.0} &\cellcolor[HTML]{e2f6cc}\textbf{500.0} &\cellcolor[HTML]{e2f6cc}\textbf{500.0} \\ \cmidrule{5-10}
\multirow{3}{*}{0.3-0.0-0.3-0.4} &\multirow{3}{*}{500} &\multirow{3}{*}{3000} &1 &\cellcolor[HTML]{e2f6cc}\textbf{353.7} &\cellcolor[HTML]{e2f6cc}\textbf{353.1} &432.9 &\cellcolor[HTML]{e2f6cc}\textbf{434.1} &489.0 &\cellcolor[HTML]{e2f6cc}\textbf{498.8} \\
& & &2 &\cellcolor[HTML]{e2f6cc}\textbf{413.0} &\cellcolor[HTML]{e2f6cc}\textbf{413.0} &453.0 &\cellcolor[HTML]{e2f6cc}\textbf{454.0} &489.0 &\cellcolor[HTML]{e2f6cc}\textbf{500.0} \\
& & &3 &\cellcolor[HTML]{e2f6cc}\textbf{417.0} &416.9 &455.0 &\cellcolor[HTML]{e2f6cc}\textbf{456.0} &489.0 &\cellcolor[HTML]{e2f6cc}\textbf{500.0} \\ \cmidrule{5-10}
\multirow{3}{*}{0.2-0.0-0.3-0.5} &\multirow{3}{*}{500} &\multirow{3}{*}{3000} &1 &203.8 &\cellcolor[HTML]{e2f6cc}\textbf{204.4} &287.6 &\cellcolor[HTML]{e2f6cc}\textbf{291.4} &373.0 &\cellcolor[HTML]{e2f6cc}\textbf{380.0} \\
& & &2 &247.0 &\cellcolor[HTML]{e2f6cc}\textbf{252.6} &316.0 &\cellcolor[HTML]{e2f6cc}\textbf{323.8} &386.0 &\cellcolor[HTML]{e2f6cc}\textbf{394.0} \\
& & &3 &247.0 &\cellcolor[HTML]{e2f6cc}\textbf{254.9} &316.0 &\cellcolor[HTML]{e2f6cc}\textbf{323.8} &386.0 &\cellcolor[HTML]{e2f6cc}\textbf{394.0} \\ \cmidrule{5-10}
\multirow{3}{*}{0.4-0.15-0.15-0.3} &\multirow{3}{*}{1000} &\multirow{3}{*}{8000} &1 &446.6 &\cellcolor[HTML]{e2f6cc}\textbf{447.4} &679.4 &\cellcolor[HTML]{e2f6cc}\textbf{680.5} &907.6 &\cellcolor[HTML]{e2f6cc}\textbf{915.9} \\
& & &2 &982.0 &\cellcolor[HTML]{e2f6cc}\textbf{985.0} &996.0 &\cellcolor[HTML]{e2f6cc}\textbf{1,000.0} &996.0 &\cellcolor[HTML]{e2f6cc}\textbf{1,000.0} \\
& & &3 &\cellcolor[HTML]{e2f6cc}\textbf{996.0} &\cellcolor[HTML]{e2f6cc}\textbf{996.0} &996.0 &\cellcolor[HTML]{e2f6cc}\textbf{1,000.0} &996.0 &\cellcolor[HTML]{e2f6cc}\textbf{1,000.0} \\ \cmidrule{5-10}
\multirow{3}{*}{0.3-0.0-0.3-0.4} &\multirow{3}{*}{1000} &\multirow{3}{*}{8000} &1 &604.0 &\cellcolor[HTML]{e2f6cc}\textbf{604.6} &773.8 &\cellcolor[HTML]{e2f6cc}\textbf{774.7} &908.7 &\cellcolor[HTML]{e2f6cc}\textbf{909.2} \\
& & &2 &\cellcolor[HTML]{e2f6cc}\textbf{808.8} &808.0 &\cellcolor[HTML]{e2f6cc}\textbf{880.0} &879.6 &948.8 &\cellcolor[HTML]{e2f6cc}\textbf{949.8} \\
& & &3 &\cellcolor[HTML]{e2f6cc}\textbf{824.3} &\cellcolor[HTML]{e2f6cc}\textbf{824.8} &888.6 &\cellcolor[HTML]{e2f6cc}\textbf{891.0} &955.0 &\cellcolor[HTML]{e2f6cc}\textbf{955.9} \\ \cmidrule{5-10}
\multirow{3}{*}{0.2-0.0-0.3-0.5} &\multirow{3}{*}{1000} &\multirow{3}{*}{8000} &1 &296.0 &\cellcolor[HTML]{e2f6cc}\textbf{296.4} &438.0 &\cellcolor[HTML]{e2f6cc}\textbf{441.3} &596.1 &\cellcolor[HTML]{e2f6cc}\textbf{612.2} \\
& & &2 &390.4 &\cellcolor[HTML]{e2f6cc}\textbf{404.7} &506.7 &\cellcolor[HTML]{e2f6cc}\textbf{523.2} &630.0 &\cellcolor[HTML]{e2f6cc}\textbf{658.9} \\
& & &3 &404.1 &\cellcolor[HTML]{e2f6cc}\textbf{424.7} &511.8 &\cellcolor[HTML]{e2f6cc}\textbf{531.2} &635.7 &\cellcolor[HTML]{e2f6cc}\textbf{662.0} \\
\bottomrule
\end{tabular}
}
\end{table*}

\begin{table*}[h] 
\caption{Numerical comparison of three algorithms---\texttt{BRKGA}, \texttt{BRKGA+FC} (results extracted from~\cite{9909110}), and our hybrid approach \texttt{BRKGA+LLM}---on a total of four real-world social network instances. For each network, the algorithms were applied 10 times to each combination of $d \in \{1, 2, 3\}$ and $k \in \{32, 64, 128\}$. Green cells indicate the best quality metrics results---higher values are better in this maximization problem.} \label{table:real-brkga-llm}
\resizebox{\linewidth}{!}{
 \begin{tabular}{cccc|ccc|ccc|ccc}\toprule
        & & & &\multicolumn{3}{c}{\textbf{$k = 32$}} &\multicolumn{3}{c}{\textbf{$k = 64$}} &\multicolumn{3}{c}{\textbf{$k = 128$}} \\ \cmidrule{5-13}
        \textbf{Instance} &\textbf{$|V|$} &\textbf{$|E|$} &\textbf{$d$} &\texttt{BRKGA} &\texttt{BRKGA+FC} &\texttt{BRKGA+LLM} &\texttt{BRKGA} &\texttt{BRKGA+FC} &\texttt{BRKGA+LLM} &\texttt{BRKGA} &\texttt{BRKGA+FC} &\texttt{BRKGA+LLM} \\ \cmidrule{1-13}
        
        \multirow{3}{*}{\textsf{soc-hamsterster}} &\multirow{3}{*}{2426} &\multirow{3}{*}{16630} &1 &1,230.0 &962.7 &\cellcolor[HTML]{e2f6cc}\textbf{1,238.9} &1,455.3 &1,184.5 &\cellcolor[HTML]{e2f6cc}\textbf{1,478.0} &1,627.8 &1,376.5 &\cellcolor[HTML]{e2f6cc}\textbf{1,731.3} \\
        & & &2 &1,751.0 &1,682.1 &\cellcolor[HTML]{e2f6cc}\textbf{1,783.2} &1,779.6 &1,778.7 &\cellcolor[HTML]{e2f6cc}\textbf{1,892.0} &1,811.0 &1,857.0 &\cellcolor[HTML]{e2f6cc}\textbf{2,115.6} \\
        & & &3 &1,788.0 &1,799.6 &\cellcolor[HTML]{e2f6cc}\textbf{1,876.0} &1,816.8 &1,850.2 &\cellcolor[HTML]{e2f6cc}\textbf{1,947.3} &1,828.0 &1,877.4 &\cellcolor[HTML]{e2f6cc}\textbf{2,180.2} \\ \cmidrule{5-13}
        \multirow{3}{*}{\textsf{sign-bitcoinotc}} &\multirow{3}{*}{5881} &\multirow{3}{*}{35592} &1 &\cellcolor[HTML]{e2f6cc}\textbf{3,479.0} &\cellcolor[HTML]{e2f6cc}\textbf{3,479.0} &\cellcolor[HTML]{e2f6cc}\textbf{3,479.0} &4,040.3 &4,041.0 &\cellcolor[HTML]{e2f6cc}\textbf{4,054.7} &4,599.9 &4,618.0 &\cellcolor[HTML]{e2f6cc}\textbf{4,606.2} \\
        & & &2 &5,632.0 &5,632.6 &\cellcolor[HTML]{e2f6cc}\textbf{5,650.2} &5,716.4 &5,715.2 &\cellcolor[HTML]{e2f6cc}\textbf{5,752.2} &5,769.0 &5,781.2 &\cellcolor[HTML]{e2f6cc}\textbf{5,835.1} \\
        & & &3 &5,838.0 &5,838.0 &\cellcolor[HTML]{e2f6cc}\textbf{5,852.7} &5,839.0 &5,839.1 &\cellcolor[HTML]{e2f6cc}\textbf{5,863.4} &5,842.0 &5,844.0 &\cellcolor[HTML]{e2f6cc}\textbf{5,868.0} \\ \cmidrule{5-13}
        \multirow{3}{*}{\textsf{soc-advogato}} &\multirow{3}{*}{6551} &\multirow{3}{*}{51332} &1 &2,464.1 &2,469.1 &\cellcolor[HTML]{e2f6cc}\textbf{2,485.9} &2,949.8 &2,948.9 &\cellcolor[HTML]{e2f6cc}\textbf{2,952.2} &3,342.2 &3,372.3 &\cellcolor[HTML]{e2f6cc}\textbf{3,385.1} \\
        & & &2 &4,142.6 &4,132.3 &\cellcolor[HTML]{e2f6cc}\textbf{4,144.9} &4,208.7 &4,207.8 &\cellcolor[HTML]{e2f6cc}\textbf{4,223.1} &4,268.5 &4,251.1 &\cellcolor[HTML]{e2f6cc}\textbf{4,330.1} \\
        & & &3 &4,280.3 &4,275.5 &\cellcolor[HTML]{e2f6cc}\textbf{4,318.6} &4,284.4 &4,280.0 &\cellcolor[HTML]{e2f6cc}\textbf{4,359.9} &4,301.2 &4,284.0 &\cellcolor[HTML]{e2f6cc}\textbf{4,431.9} \\ \cmidrule{5-13}
        \multirow{3}{*}{\textsf{soc-wiki-elec}} &\multirow{3}{*}{7118} &\multirow{3}{*}{107071} &1 &2,167.0 &2,176.7 &\cellcolor[HTML]{e2f6cc}\textbf{2,188.0} &2,265.6 &2,268.6 &\cellcolor[HTML]{e2f6cc}\textbf{2,286.1} &2,367.7 &2,366.5 &\cellcolor[HTML]{e2f6cc}\textbf{2,408.8} \\
        & & &2 &2,354.7 &2,355.1 &\cellcolor[HTML]{e2f6cc}\textbf{2,365.0} &2,390.0 &2,388.0 &\cellcolor[HTML]{e2f6cc}\textbf{2,409.7} &2,454.5 &2,427.5 &\cellcolor[HTML]{e2f6cc}\textbf{2,478.6} \\
        & & &3 &2,357.1 &2,357.3 &\cellcolor[HTML]{e2f6cc}\textbf{2,366.2} &2,389.5 &2,389.7 &\cellcolor[HTML]{e2f6cc}\textbf{2,406.4} &2,452.2 &2,426.5 &\cellcolor[HTML]{e2f6cc}\textbf{2,474.2} \\
        \bottomrule
        
        \end{tabular}%

}
\end{table*}

In the next set of experiments, we applied the \texttt{BRKGA+FC}~\cite{9909110}, in addition to \texttt{BRKGA} and \texttt{BRKGA+LLM}, to the four larger real-world social networks. Remember that \texttt{BRKGA+FC} is a hybrid approach that uses a hand-crafted GNN approach for biasing the search process of \texttt{BRKGA}. The results are shown---again for each combination of $d \in \{1, 2, 3\}$  and $k \in \{32, 64, 128\}$---in Table~\ref{table:real-brkga-llm}. The computation time limit for the three approaches was 900 CPU seconds, as in the previously outlined experiments. Moreover, the numbers in the tables are averages over 10 algorithm runs. The results show that \texttt{BRKGA+LLM} outperforms both approaches, with higher margins than those observed in the context of smaller synthetic networks. This holds especially for a growing value of $k$. This is interesting as the prompts only contained an example solution for $k=32$. This indicates the LLM's ability to uncover meaningful patterns in the example graph, respectively, in the solution provided in the prompts. \\


 Finally, we aimed to test how meaningful the LLM output really is. For this purpose, we produced two additional variants of \texttt{BRKGA+LLM}: the one called \texttt{static} is obtained by replacing the LLM output with probabilities obtained by random alpha and beta values. The second one, called \texttt{dynamic}, is a similar variant in which the LLM output is replaced with probabilities re-computed at each iteration based on newly determined random alpha and beta values. All three algorithm variants were applied 10 times to each of the four real-world social networks for each combination of $d \in \{1, 2, 3\}$  and $k \in \{32, 64, 128\}$. The results averaged over the 10 runs are shown in Table~\ref{table:random-llm}. They clearly indicate that guidance by the LLM output is much more useful than random guidance. \\
 

\begin{table*}[!t] 
\caption{Comparison of the LLM output with random values. \texttt{static} refers to a variant of \texttt{BRKGA+LLM} in which the LLM output is replaced by probabilities computed based on random alpha and beta values. \texttt{dynamic} refers to a very similar \texttt{BRKGA+LLM} variant in which the random values for the alpha's and beta's are dynamically changed at each iteration. Green cells indicate the best quality metrics results---higher values are better in this maximization problem.} \label{table:random-llm}
\resizebox{\linewidth}{!}{
\begin{tabular}{cccc|ccc|ccc|ccc}\toprule

& & & &\multicolumn{3}{c}{\textbf{$k = 32$}} &\multicolumn{3}{c}{\textbf{$k = 64$}} &\multicolumn{3}{c}{\textbf{$k = 128$}} \\\cmidrule{5-13}
\textbf{Instance} &\textbf{$|V|$} &\textbf{$|E|$} &\textbf{$d$} &\texttt{static} &\texttt{dynamic} &\texttt{BRKGA+LLM} &\texttt{static} &\texttt{dynamic} &\texttt{BRKGA+LLM} &\texttt{static} &\texttt{dynamic} &\texttt{BRKGA+LLM} \\ \cmidrule{1-13}

\multirow{3}{*}{\textsf{soc-hamsterster}} &\multirow{3}{*}{2426} &\multirow{3}{*}{16630} &1 &1,226.5 &1,227.1 &\cellcolor[HTML]{e2f6cc}\textbf{1,238.9} &1,419.7 &1,418.5 &\cellcolor[HTML]{e2f6cc}\textbf{1,500.2} &1,605.6 &1,609.0 &\cellcolor[HTML]{e2f6cc}\textbf{1,791.5} \\
& & &2 &1,744.8 &1,746.3 &\cellcolor[HTML]{e2f6cc}\textbf{1,783.2} &1,777.5 &1,781.3 &\cellcolor[HTML]{e2f6cc}\textbf{1,972.5} &1,811.0 &1,811.0 &\cellcolor[HTML]{e2f6cc}\textbf{2,150.2} \\
& & &3 &1,788.0 &1,788.0 &\cellcolor[HTML]{e2f6cc}\textbf{1,876.0} &1,816.0 &1,811.8 &\cellcolor[HTML]{e2f6cc}\textbf{2,056.6} &1,825.2 &1,822.4 &\cellcolor[HTML]{e2f6cc}\textbf{2,211.0} \\ \cmidrule{5-13}
\multirow{3}{*}{\textsf{sign-bitcoinotc}} &\multirow{3}{*}{5881} &\multirow{3}{*}{35592} &1 &\cellcolor[HTML]{e2f6cc}\textbf{3,479.0} &\cellcolor[HTML]{e2f6cc}\textbf{3,479.0} &\cellcolor[HTML]{e2f6cc}\textbf{3,479.0} &4,037.6 &4,038.5 &\cellcolor[HTML]{e2f6cc}\textbf{4,061.0} &4,593.9 &4,594.1 &\cellcolor[HTML]{e2f6cc}\textbf{4,628.2} \\
& & &2 &5,632.1 &5,632.1 &\cellcolor[HTML]{e2f6cc}\textbf{5,650.2} &5,715.3 &5,715.3 &\cellcolor[HTML]{e2f6cc}\textbf{5,767.7} &5,761.8 &5,734.4 &\cellcolor[HTML]{e2f6cc}\textbf{5,842.0} \\
& & &3 &5,838.0 &5,838.0 &\cellcolor[HTML]{e2f6cc}\textbf{5,852.7} &5,839.0 &5,839.0 &\cellcolor[HTML]{e2f6cc}\textbf{5,873.9} &5,842.0 &5,842.0 &\cellcolor[HTML]{e2f6cc}\textbf{5,874.0} \\ \cmidrule{5-13}
\multirow{3}{*}{\textsf{soc-advogato}} &\multirow{3}{*}{6551} &\multirow{3}{*}{51332} &1 &2,463.7 &2,464.1 &\cellcolor[HTML]{e2f6cc}\textbf{2,485.9} &2,949.1 &2,949.4 &\cellcolor[HTML]{e2f6cc}\textbf{2,958.5} &3,338.5 &3,339.2 &\cellcolor[HTML]{e2f6cc}\textbf{3,402.0} \\
& & &2 &4,141.1 &4,138.7 &\cellcolor[HTML]{e2f6cc}\textbf{4,144.9} &4,207.2 &4,206.1 &\cellcolor[HTML]{e2f6cc}\textbf{4,234.8} &4,267.2 &4,266.4 &\cellcolor[HTML]{e2f6cc}\textbf{4,357.3} \\
& & &3 &4,279.0 &4,276.6 &\cellcolor[HTML]{e2f6cc}\textbf{4,318.6} &4,281.3 &4,283.4 &\cellcolor[HTML]{e2f6cc}\textbf{4,377.8} &4,301.1 &4,299.7 &\cellcolor[HTML]{e2f6cc}\textbf{4,451.5} \\ \cmidrule{5-13}
\multirow{3}{*}{\textsf{soc-wiki-elec}} &\multirow{3}{*}{7118} &\multirow{3}{*}{107071} &1 &2,166.6 &2,169.1 &\cellcolor[HTML]{e2f6cc}\textbf{2,188.0} &2,265.0 &2,264.7 &\cellcolor[HTML]{e2f6cc}\textbf{2,295.0} &2,367.1 &2,366.5 &\cellcolor[HTML]{e2f6cc}\textbf{2,417.1} \\
& & &2 &2,354.8 &2,354.8 &\cellcolor[HTML]{e2f6cc}\textbf{2,365.0} &2,389.0 &2,389.8 &\cellcolor[HTML]{e2f6cc}\textbf{2,418.6} &2,454.4 &2,454.8 &\cellcolor[HTML]{e2f6cc}\textbf{2,484.0} \\
& & &3 &2,357.3 &2,357.1 &\cellcolor[HTML]{e2f6cc}\textbf{2,366.2} &2,389.8 &2,390.1 &\cellcolor[HTML]{e2f6cc}\textbf{2,419.6} &2,451.9 &2,451.0 &\cellcolor[HTML]{e2f6cc}\textbf{2,485.0} \\
\bottomrule

\end{tabular}

}
\end{table*}

We can conclude that guidance by LLM output clearly leads to an improved algorithm and results. In other words, these experiments demonstrate that an LLM can provide valuable information to inform a metaheuristic. In fact, we would expect an even higher benefit through LLM guidance in the context of even larger networks. However, this is currently not possible due to the reasons outlined before. 


\subsubsection{Dimension 2 of the Evaluation Framework: Alternative Techniques for Guiding MHs}


To assess the reliability of the LLM output in an alternative way, we decided to compare \texttt{BRKGA+LLM} to an algorithm variant in which the alpha and beta values are obtained by an explicit parameter tuning procedure using the \texttt{irace} tool~\cite{lopez2016irace}. This algorithm variant will henceforth be called \texttt{BRKGA+irace}. In particular, for each of the four large social networks---mentioned in Section~\ref{subsection:dataset-restriction} and already used, for example, in Table~\ref{table:random-llm}---we applied a tuning procedure for obtaining well-working alpha and beta values in the following way. First, for every combination of $d \in \{1, 2, 3\}$  and $k \in \{32, 64, 128\}$ a training instance was generated. Second, \texttt{irace} was applied with a budget of \num{1000} algorithm runs, using (as before) a computation time limit of 900 CPU seconds per run. After obtaining the final alpha and beta values from \texttt{irace} for each network, \texttt{BRKGA+irace} was applied under the same conditions as \texttt{BRKGA+LLM} to each of the four problem instances. The results are shown in Table~\ref{table:irace-llm}. We observe that while \texttt{BRKGA+irace} outperforms \texttt{BRKGA+LLM} in the \textsf{soc-hamsterster} and \textsf{soc-wiki-elec} instances, the opposite is the case for the \textsf{sign-bitcoinotc} instance. In the \textsf{soc-advogato} instance, performance is similar except for $d=3$, where \texttt{BRKGA+irace} shows better results. For putting these results into perspective, consider that the alpha and beta values of \texttt{BRKGA+irace} were obtained through a specific tuning process that required significant computational time—approximately 59460 minutes (single-thread) or 2696.7 minutes (parallel jobs on SLURM) for the sum of the four problem instances (see Table \ref{table:cpu-seconds-approach}).\footnote{The irace logs can be found in the \texttt{supplementary material/} folder in the repository.} In contrast, the LLM output (Claude-3-Opus) is obtained in just 5.3 minutes via an API call over HTTP. While this is not a fully fair comparison due to the different hardware setups, the substantial difference in runtime highlights the efficiency of the LLM-based approach. In some scenarios, it may be preferable to pay for API access rather than maintain dedicated servers that require significantly more time to achieve comparable results. 

\begin{table}[!t]\centering
\caption{Comparison of computational time (in seconds) for \texttt{irace} (single-threaded and parallel on SLURM) vs. LLM (Claude-3-Opus) via API, highlighting significant time differences despite hardware variations.}\label{table:cpu-seconds-approach}
\scriptsize
\resizebox{\linewidth}{!}{
\begin{tabular}{lcc|cc}\toprule
&\multicolumn{3}{c}{Time (seconds)} \\\cmidrule{2-4}
&\multicolumn{2}{c}{irace} &LLM \\\cmidrule{2-4}
Instance &Single-Thread (Serial) &Multiple Jobs in SLURM (Parallel) &Anthropic API (HTTP) \\\midrule
soc-wiki-elec &891900 &46422 &119 \\
soc-advogato &891900 &40002 &114 \\
sign-bitcoinotc &891900 &37292 &51 \\
soc-hamsterster &891900 &38087 &36 \\
\hline \hline
Total time (minutes) &59460.0 &2696.7 &5.3 \\
\end{tabular}
}
\end{table}

We used Pearson's correlation coefficient~\cite{Benesty2009} to identify relationships between the alpha and beta parameters obtained from \texttt{irace} and those from the LLM output (see the rows labeled with $\rho_{\text{irace}, \text{LLM}}$ in Table~\ref{table:comparison-alpha-beta}). A value close to $-1$ indicates a negative correlation, meaning that when one value increases, the other decreases. A value close to $+1$ indicates a positive correlation, where both values tend to move together. A value close to $0$ means there is no clear relationship between the two series. Our observations are as follows:

\begin{enumerate}
    \item In the instances where \texttt{BRKGA+irace} clearly dominates---\textsf{soc-hamsterster} and \textsf{soc-wiki-elec}---there is a moderate to strong negative correlation concerning the alpha values. However, concerning the beta values, there is no clear relationship in the context of \textsf{soc-hamsterster}, whereas there is a negative correlation in the context of \textsf{soc-wiki-elec}. Thus, negative correlations are predominant. This suggests that the values determined by \texttt{irace} and the LLM tend to move in opposite directions, and based on the results, the direction chosen by \texttt{irace} appears to be the better one.

    \item In contrast, in the \textsf{sign-bitcoinotc} instance, where \texttt{BRKGA+LLM} outperforms \texttt{BRKGA+irace}, there is no correlation between the two methods concerning the alpha values. The LLM's prediction is unique and unrelated to \texttt{irace}'s, demonstrating that the LLM could find good results that \texttt{irace} missed.
    
    \item The results for the \textsf{soc-advogato} instance are inconclusive since, for each $k \in \{32, 64, 128\}$, there is no definitive winner between \texttt{BRKGA+irace} and \texttt{BRKGA+LLM}. A negative correlation concerning the alpha values but a positive correlation in the beta values can be observed. This suggests that either \texttt{irace} has identified the alpha values correctly but not the beta values, or the LLM has identified the beta values correctly but not the alpha values.
\end{enumerate}

\begin{table*}[!t] 
\caption{A numerical comparison of \texttt{BRKGA+LLM} and \texttt{BRKGA+irace}. In the latter algorithm, the alpha and beta values are determined by tuning through \texttt{irace}. Green cells indicate the best quality metrics results---higher values are better in this maximization problem.} \label{table:irace-llm}
    
\resizebox{\linewidth}{!}{
     \begin{tabular}{cccc|cc|cc|cc}\toprule
        & & & &\multicolumn{2}{c}{\textbf{$k = 32$}} &\multicolumn{2}{c}{\textbf{$k = 64$}} &\multicolumn{2}{c}{\textbf{$k = 128$}} \\\cmidrule{5-10}
        \textbf{Instance} &\textbf{$|V|$} &\textbf{$|E|$} &\textbf{$d$} &\texttt{BRKGA+irace}&\texttt{BRKGA+LLM} &\texttt{BRKGA+irace} &\texttt{BRKGA+LLM}  
        &\texttt{BRKGA+irace} &\texttt{BRKGA+LLM} \\\midrule
        
        \multirow{3}{*}{\textsf{soc-hamsterster}} &\multirow{3}{*}{2426} &\multirow{3}{*}{16630} &1 &\cellcolor[HTML]{e2f6cc}\textbf{1,242.4} &1,238.9 &\cellcolor[HTML]{e2f6cc}\textbf{1,487.9} &1,478.0 &\cellcolor[HTML]{e2f6cc}\textbf{1,763.2} &1,731.3 \\
        & & &2 &\cellcolor[HTML]{e2f6cc}\textbf{1,816.3} &1,783.2 &\cellcolor[HTML]{e2f6cc}\textbf{1,956.9} &1,892.0 &\cellcolor[HTML]{e2f6cc}\textbf{2,128.7} &2,115.6 \\
        & & &3 &\cellcolor[HTML]{e2f6cc}\textbf{1,931.4} &1,876.0 &\cellcolor[HTML]{e2f6cc}\textbf{2,038.4} &1,947.3 &\cellcolor[HTML]{e2f6cc}\textbf{2,192.5} &2,180.2 \\ \cmidrule{5-10}
        \multirow{3}{*}{\textsf{sign-bitcoinotc}} &\multirow{3}{*}{5881} &\multirow{3}{*}{35592} &1 &3,478.8 &\cellcolor[HTML]{e2f6cc}\textbf{3,479.0} &4,054.5 &\cellcolor[HTML]{e2f6cc}\textbf{4,054.7} &4,606.0 &\cellcolor[HTML]{e2f6cc}\textbf{4,606.2} \\
        & & &2 &5,642.3 &\cellcolor[HTML]{e2f6cc}\textbf{5,650.2} &5,749.1 &\cellcolor[HTML]{e2f6cc}\textbf{5,752.2} &5,823.5 &\cellcolor[HTML]{e2f6cc}\textbf{5,835.1} \\
        & & &3 &5,851.6 &\cellcolor[HTML]{e2f6cc}\textbf{5,852.7} &5,860.4 &\cellcolor[HTML]{e2f6cc}\textbf{5,863.4} &5,867.1 &\cellcolor[HTML]{e2f6cc}\textbf{5,868.0} \\ \cmidrule{5-10}
        \multirow{3}{*}{\textsf{soc-advogato}} &\multirow{3}{*}{6551} &\multirow{3}{*}{51332} &1 &\cellcolor[HTML]{e2f6cc}\textbf{2,487.3} &2,485.9 &2,951.9 &\cellcolor[HTML]{e2f6cc}\textbf{2,952.2} &3,380.9 &\cellcolor[HTML]{e2f6cc}\textbf{3,385.1} \\
        & & &2 &4,141.3 &\cellcolor[HTML]{e2f6cc}\textbf{4,144.9} &\cellcolor[HTML]{e2f6cc}\textbf{4,224.6} &4,223.1 &4,329.2 &\cellcolor[HTML]{e2f6cc}\textbf{4,330.1} \\
        & & &3 &\cellcolor[HTML]{e2f6cc}\textbf{4,320.1} &4,318.6 &\cellcolor[HTML]{e2f6cc}\textbf{4,366.8} &4,359.9 &\cellcolor[HTML]{e2f6cc}\textbf{4,441.0} &4,431.9 \\ \cmidrule{5-10}
        \multirow{3}{*}{\textsf{soc-wiki-elec}} &\multirow{3}{*}{7118} &\multirow{3}{*}{107071} &1 &\cellcolor[HTML]{e2f6cc}\textbf{2,188.7} &2,188.0 &\cellcolor[HTML]{e2f6cc}\textbf{2,293.2} &2,286.1 &\cellcolor[HTML]{e2f6cc}\textbf{2,411.5} &2,408.8 \\
        & & &2 &\cellcolor[HTML]{e2f6cc}\textbf{2,375.3} &2,365.0 &\cellcolor[HTML]{e2f6cc}\textbf{2,417.5} &2,409.7 &\cellcolor[HTML]{e2f6cc}\textbf{2,482.90} &2,478.6 \\
        & & &3 &\cellcolor[HTML]{e2f6cc}\textbf{2,378.8} &2,366.2 &\cellcolor[HTML]{e2f6cc}\textbf{2,415.3} &2,406.4 &\cellcolor[HTML]{e2f6cc}\textbf{2,478.60} &2,474.2 \\
        \bottomrule

\end{tabular}
}
\end{table*}

\begin{table*}[!t] 
\caption{The alpha and beta values as determined by \texttt{irace} and the \texttt{LLM} for each case. Pearson's correlation coefficient ($\rho_{\text{irace}, \text{LLM}}$) is used to quantify the relationships between the two sets of values.} \label{table:comparison-alpha-beta}
\centering
        \resizebox{0.7\linewidth}{!}{
        \begin{tabular}{r|cc|cc|cc|ccc}\toprule
        &\multicolumn{2}{c}{\textsf{soc-hamsterster}} &\multicolumn{2}{c}{\textsf{sign-bitcoinotc}} &\multicolumn{2}{c}{\textsf{soc-advogato}} &\multicolumn{2}{c}{\textsf{soc-wiki-elec}} \\\cmidrule{2-9}
        &\texttt{irace} &\texttt{LLM} &\texttt{irace} &\texttt{LLM} &\texttt{irace} &LLM &\texttt{irace} &\texttt{LLM} \\\midrule
        $\alpha_1$ &0.40 &0.10 &0.28 &0.15 &0.21 &0.15 &0.08 &0.15 \\
        $\alpha_2$ &0.08 &0.30 &0.14 &0.25 &0.21 &0.25 &0.05 &0.25 \\
        $\alpha_3$ &0.03 &0.20 &0.30 &0.20 &0.12 &0.35 &0.21 &0.20 \\
        $\alpha_4$ &0.40 &0.10 &0.18 &0.30 &0.34 &0.15 &0.25 &0.30 \\
        $\alpha_5$ &0.09 &0.30 &0.10 &0.10 &0.12 &0.10 &0.41 &0.10 \\ \midrule
        $\rho_{\text{irace}, \text{LLM}}$ &\multicolumn{2}{c}{-0.85} &\multicolumn{2}{c}{0.02} &\multicolumn{2}{c}{-0.27}  &\multicolumn{2}{c}{-0.38} \\ \midrule
        $\beta_1$ &0.78 &0.60 &0.61 &0.70 &0.31 &0.60 &0.61 &0.70 \\
        $\beta_2$ &0.83 &0.60 &0.21 &0.60 &0.65 &0.70 &0.92 &0.60 \\
        $\beta_3$ &0.65 &0.90 &0.01 &0.80 &0.83 &0.90 &0.19 &0.90 \\
        $\beta_4$ &0.01 &0.60 &0.51 &0.05 &0.75 &0.60 &0.80 &0.60 \\
        $\beta_5$ &0.75 &0.60 &0.50 &0.10 &0.86 &0.70 &0.51 &0.50 \\ \midrule
        $\rho_{\text{irace}, \text{LLM}}$ &\multicolumn{2}{c}{0.08} &\multicolumn{2}{c}{-0.55} &\multicolumn{2}{c}{0.55} &\multicolumn{2}{c}{-0.67} \\
        \bottomrule
        \end{tabular}%
}
\end{table*}

We believe that, although \texttt{BRKGA+LLM's} results are not favorable compared to those obtained with the help of \texttt{irace} in the context of two out of four problem instances, our approach benefits from a significantly reduced computational effort (see Table\ref{table:cpu-seconds-approach}). This still holds when the computation time required to extract five metric values to build the prompts is taken into account. Additionally, with model improvements or prompt adjustments---for example, the LLM currently, by its own internal decision, produces values divisible by 0.05, leading to a loss of precision compared to \texttt{irace}---greater precision could enable the LLM to improve in the future.

\subsubsection{Dimension 3 of the Evaluation Framework: Prompt Quality}

In the final dimension, we transition from a numerical analysis---as conducted in the previous two dimensions---to a more interpretative approach, where we discuss the complex process of designing a well-working prompt tailored to the considered optimization problem. To accomplish this, we first examine the five selected metrics already introduced in Section~\ref{sec:prompt-engineering}. This is crucial because every additional metric enlarges the prompt. And the larger the prompt, the smaller the maximum network size that---due to the limited size of the context window---can be passed to the LLM. Also, larger prompts are associated with increased financial costs. Hence, it is essential to determine whether similar or even superior results can be achieved using less information. To investigate this, we perform two experiments: the first assesses the correlation between each pair of metrics, and the second removes information from the prompt to analyze the impact on the LLM results. \\

\textbf{Correlation between metrics.} For this analysis, we utilize a matrix of plots provided in Figure~\ref{fig:metric} concerning the \textsf{soc-hamsterster} instance for which \texttt{BRKGA+LLM} significantly outperformed \texttt{BRKGA}; see Table~\ref{table:real-brkga-llm}. Hereby, the plots in the upper triangle of the matrix are scatter plots that display the values of all pairs of metrics. For example, the second plot in the first matrix row shows the scatter plot concerning the values of metrics \texttt{out-degree} (x-axis) and \texttt{in-degree} (y-axis). In contrast, the plots in the lower triangle are kernel density estimation (KDE) plots for each pair of metrics. KDE plots provide a smoothed representation of the underlying data distribution, aiding in the identification of patterns such as clusters, outliers, and non-linear relationships. Lastly, the diagonal showcases univariate KDE plots for each variable, analogous to a histogram, depicting the distribution of each variable independently. Note that a corresponding graphic concerning the other problem instances is provided in the GitHub repository, whose link can be found in Section~\ref{subsec:repository}. The following observations can be made based on Figure~\ref{fig:metric}:
\begin{itemize}
    \item The upper triangle reveals that all pairs of metric comparisons exhibit a non-linear pattern, albeit to varying degrees. For instance, the relationship between \texttt{in-degree} and  \texttt{closeness} indicates a complex interaction between the metrics, suggesting that each metric contributes additional information about the problem. This finding indicates that none of the considered metrics is superfluous. 
    \item The KDE plots in the lower triangle highlight non-linear relationships that may not be as apparent in the scatter plots. The relationship between \texttt{pagerank} and all other metrics (bottom row) indicates mainly a more linear relationship with other metrics, which was not as evident in the scatter plots. Still, the outliers shown in the scatter plots do not allow \texttt{pagerank} to be excluded from the prompt.
    \item  Finally, the diagonal of univariate KDE plots reveals that certain metrics, such as \texttt{betweenness}, contain more frequently occurring values---observe the peak in the plot. In other words, some \textsf{betweenness} values repeat across many nodes in this instance. This insight suggests developing a prompt design strategy to further decrease the prompt size.
\end{itemize}

Upon examining each metric, we conclude that all are potentially relevant and contribute to the outcome. However, it is still possible that adjusting the set of metrics by removing certain metrics or adding alternative ones could lead to even better results.\footnote{Financial limitations prevent from conducting extensive experiments and from studying every aspect of the prompt, especially for large-size, challenging scenarios; hence, careful consideration is essential.} \\

\Figure[h!](topskip=0pt, botskip=0pt, midskip=0pt)[width=1\linewidth]{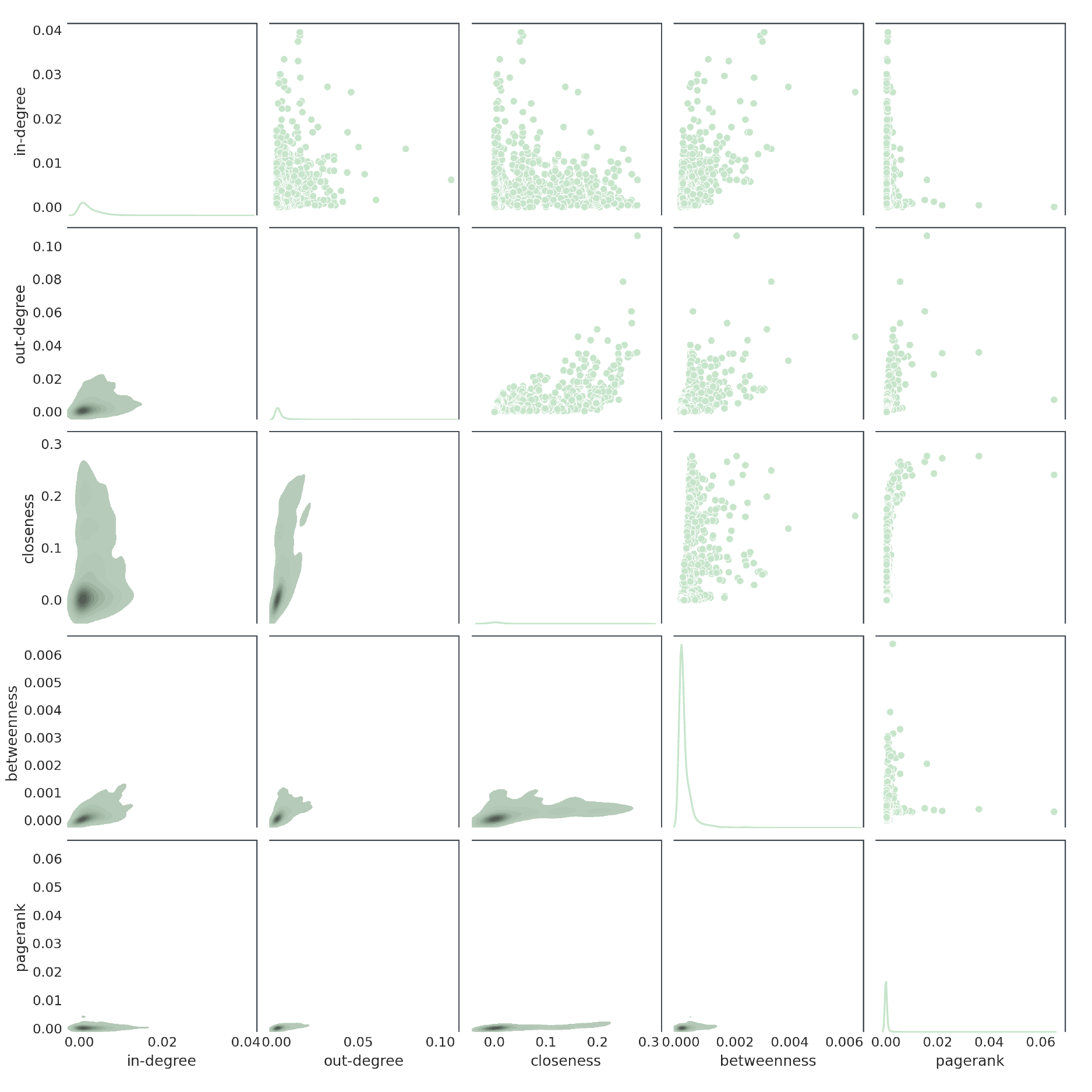}
{Correlations between all pairs of the five considered metrics concerning the \textsf{soc-hamsterster} network.\label{fig:metric}}

\textbf{Removal of information.} During our investigation, we discovered that modifying the prompt has the following effects:

\begin{itemize}
    \item When removing the graph metrics from the prompt (by removing the \textsf{[EXAMPLE GRAPH]} tag), the LLM is still capable of generating useful response values. This implies that the LLM's extensive pre-training enables it to infer alpha and beta values, leveraging its prior knowledge of relevant metrics for social network node coverage problems. Despite this prior knowledge, providing an example graph (in terms of the five metric values per node) allows the LLM to refine the alpha and beta values to better suit the considered $k$-$d$DSP problem, resulting in improved output.
    \item Expressing metric values for each node in scientific numerical notation does not compromise the quality of the LLM's response (\textsf{[DATA]} tag). This approach offers a dual benefit: it enables greater precision while reducing the character count and, therefore, the number of tokens in the prompt.
    \item The beta values play a crucial role in shaping the response quality. That is, omitting them significantly reduces the quality of the LLM's output. By assigning importance weights to each metric (alpha values) and requesting an expected value (beta), we apparently enable the LLM to uncover more subtle patterns in the evaluation graph's metrics (\textsf{[EVALUATION GRAPH]} tag), ultimately leading to enhanced results.
\end{itemize}

To summarize, while the LLM possesses prior knowledge, it appears insufficient to enable the model to independently identify patterns in tabular numerical data. \\

\textbf{Differences in node selection.} Finally, we wanted to study how the use of LLM output leads to the selection of different nodes for solutions produced by \texttt{BRKGA+LLM} in comparison to \texttt{BRKGA}. For this purpose, Figure~\ref{fig:analysis-metric} shows the node probabilities computed based on the alpha and beta values (black line) in relation to the (normalized) values of the five metrics, exemplary in the context of the synthetic graph \textsf{0.2-0.0-0.3-0.5}. The x-axis ranges over all 500 graph nodes, ordered by a non-increasing LLM-probability. Moreover, by means of horizontal lines the graphic marks the nodes chosen for the best \texttt{BRKGA} solution (dotted), the best \texttt{BRKGA+LLM} solution (solid), and their intersection (dashed). In the following, we point out three specific cases highlighted as (a), (b), and (c) in Figure~\ref{fig:analysis-metric}:

\begin{enumerate}
    \item One of the nodes selected by \texttt{BRKGA+LLM} (second solid green line) has a much higher \textsf{closeness} metric value than it has an \textsf{out-degree} metric value. Remember that \textsf{out-degree} is the standard metric used by \texttt{BRKGA}. This indicates that the LLM can identify more suitable nodes by blending information from several available metrics.
    \item Similar to (a), it can be observed that for the first and last node selected by \texttt{BRKGA+LLM}, the \textsf{closeness} value is higher than the one of \textsf{out-degree}. In contrast, in the context of those nodes that are shared by both \texttt{BRKGA} and \texttt{BRKGA+LLM} (green dashed lines), \textsf{closeness} is high, but so is \textsf{out-degree}.
    \item Cases in which \textsf{closeness} is high and \textsf{out-degree} is relatively lower in the context of nodes selected by \texttt{BRKGA+LLM} can also be seen in this example. This indicates that, for the \textsf{0.2-0.0-0.3-0.5} network, the best solution is achieved due to the LLM's ability to recognize the importance of \textsf{closeness} for certain nodes, an importance that pure \texttt{BRKGA} cannot detect due to a lack of information.
\end{enumerate}


\Figure[h!](topskip=0pt, botskip=0pt, midskip=0pt)[width=1\linewidth]{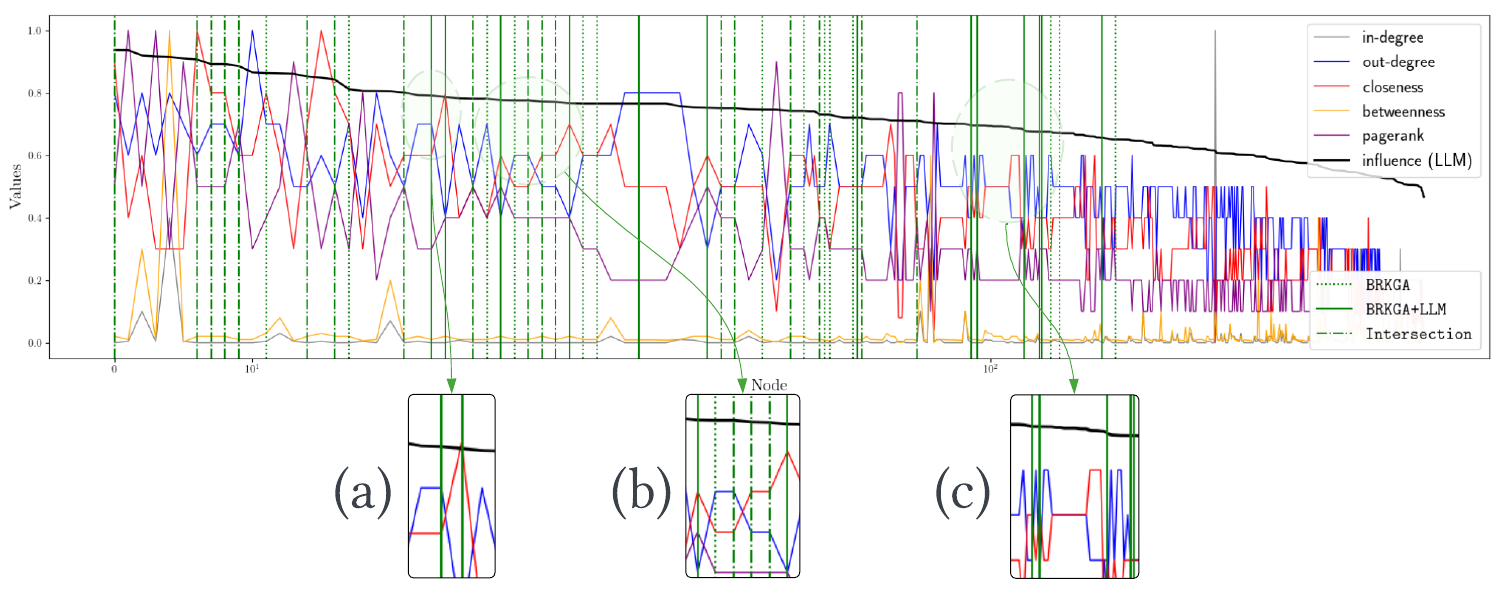}
{Analysis of the probabilities computed based on the alpha and beta values (black line) in relation to the (normalized) values of the five metrics. The x-axis ranges of all 500 nodes of the synthetic graph \textsf{0.2-0.0-0.3-0.5} ordered by a non-increasing LLM-probability. Moreover, the graphic marks the nodes chosen for the best \texttt{BRKGA} solution, the best \texttt{BRKGA+LLM} solution, and their intersection.\label{fig:analysis-metric}}

\subsection{Visual Comparative Analysis}\label{emp:visual}

Numerical analysis can fall short of capturing the full complexity of a metaheuristics' search process due to its stochastic nature. Visual tools have emerged in recent years to address this limitation. They were developed to provide a more comprehensive understanding and additional insight. One such tool is STNWeb~\cite{CHACONSARTORI2023100558}, which generates directed graphs from algorithm trajectories to visualize how these algorithms navigate the search space. This allows to compare and justify the performance of different algorithms. A visual analysis is presented in this section to understand better the advantages of our \texttt{BRKGA+LLM} approach over \texttt{BRKGA} and \texttt{BRKGA+FC}.

\Figure[h](topskip=0pt, botskip=0pt, midskip=0pt)[width=1\linewidth]{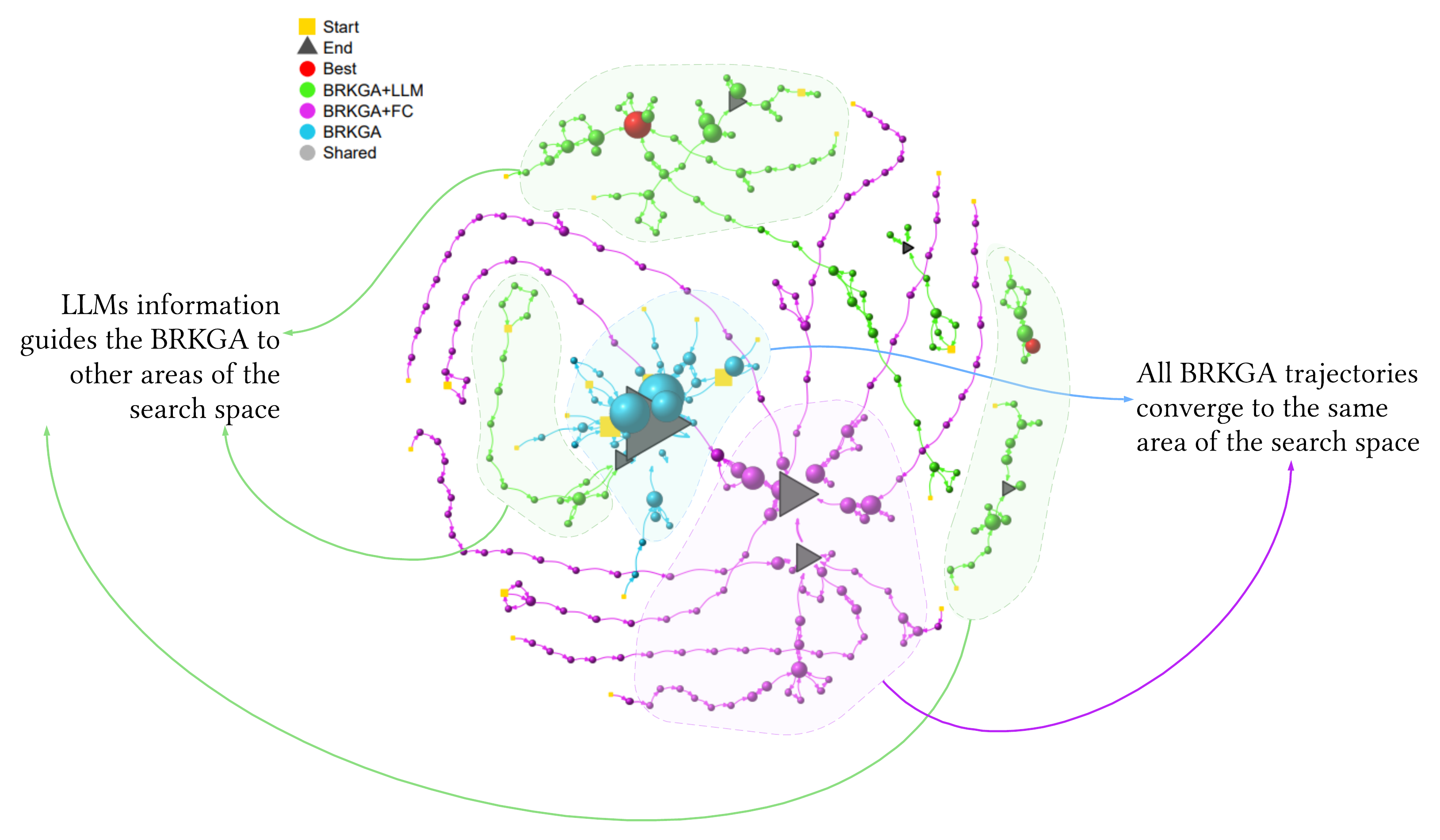}
{STNWeb-generated plot comparing the trajectories of \texttt{BRKGA} (cyan), \texttt{BRKGA+FC} (magenta) and \texttt{BRKGA+LLM} (green) over 10 runs on the \textsf{soc-hamsterster} instance (with $d=1$ and $k=32$). This plot was generated using the so-called \textsf{agglomerative clustering partitioning} method available in STNWeb, with the number of clusters set to approximately 20\% of the total, allowing for a visualization focusing on the essential characteristics.\label{fig:stn}}

Figure~\ref{fig:stn} shows a STNWeb plot displaying 10 runs of each \texttt{BRKGA+LLM}, \texttt{BRKGA}, and \texttt{BRKGA+FC} when applied to the \textsf{soc-hamsterster} instance. The following analysis highlights the insights that can be obtained from this visualization. However, first of all, let us explain the technical elements of the plot:

\begin{enumerate}
    \item Each of the 30 algorithm runs is shown as a trajectory---that is, a directed path---in the search space. Hereby, the trajectories of the three algorithms are distinguished by color: \texttt{BRKGA} (cyan~\tikz\draw[cyan,fill=cyan] (0,0) circle (.5ex);), \texttt{BRKGA+FC} (magenta~\tikz\draw[magenta,fill=magenta] (0,0) circle (.5ex);), and \texttt{BRKGA+LLM} (green~\tikz\draw[green,fill=green] (0,0) circle (.5ex);).
    \item Trajectory starting points are marked by a yellow square~(\tikz\draw[yellow,fill=yellow] (0,0) rectangle (.9ex, .9ex);). Moreover, trajectory end points are generally marked by a black triangle~(\begin{tikzpicture}\node[buffer]{Test};\end{tikzpicture}).
    \item A trajectory consists of multiple solutions (nodes,~\tikz\draw[cyan,fill=cyan] (0,0) circle (.5ex);, ~\tikz\draw[magenta,fill=magenta] (0,0) circle (.5ex);, and~\tikz\draw[green,fill=green] (0,0) circle (.5ex);) connected by directed edges, each with an associated fitness value. Since this is a maximization problem ($k$-$d$DSP), the fitness increases as it approaches the end of the trajectory~\begin{tikzpicture}\node[buffer]{Test};\end{tikzpicture}.
    \item The size of each node indicates the number of trajectories that have passed through it. 
    \item A red node~\tikz\draw[red,fill=red] (0,0) circle (.5ex); includes a best solution found by all 30 algorithm trajectories. Note that different best solutions might have been found.
\end{enumerate}

Several interesting observations can be made based on Figure~\ref{fig:stn}. First of all, only \texttt{BRKGA+LLM} can find best solutions (see the two red dots). Moreover, the two best solutions found by \texttt{BRKGA+LLM} are rather different from each other, as they are found in different areas of the search space. The three algorithms seem to be attracted by different areas of the search space. Moreover, while \texttt{BRKGA} and \texttt{BRKGA+FC} clearly converge to solutions that are close to each other, this is not so much the case for \texttt{BRKGA+LLM}, which does not show a clear convergence behavior towards a single area of the search space. Finally, note that the search trajectories of \texttt{BRKGA} are much shorter than those of the two hybrid approaches. Additional STN plots concerning other problem instances can be found in the repository whose link was provided in Section~\ref{subsec:repository}. \\


Following the empirical study of our prompt's design and quality (Section~\ref{emp:analysis}) and following a visual analysis, we can conclude that our proposed hybridization achieves its intended objective: showing that LLMs can generate heuristic information that can be used to improve the search process of a metaheuristic. Interestingly, our approach has also outperformed an alternative hybridization scheme that used heuristic information produced by a hand-crafted, specifically trained, graph neural network from~\cite{9909110}. However, successfully integrating LLMs into MHs involves addressing several critical issues and open questions, detailed in the next section.

\section{Discussion and open questions}\label{sec:discussion}

The LLM frenzy continues to gain momentum, with new papers appearing daily praising their virtues. LLMs are being applied far and wide, from tackling complex problems to simplifying mundane tasks. While it is uncertain whether their utility is universally applicable, our approach reveals the potential of LLMs to serve as pattern recognizers, uncovering hidden patterns and providing researchers with insights to boost their optimization algorithms. Our study gives rise to several open questions, including the following ones:
\begin{itemize}
    \item \textbf{Are LLMs merely stochastic parrots or black boxes capable of reasoning?} In the influential paper by Bender et al.~\cite{10.1145/3442188.3445922}, the authors raise concerns about the risks of using LLMs and question their necessity. They argue that LLMs are trained on vast amounts of data based on probability distributions but lack any reference to meaning, earning them the label of \textit{stochastic parrots}. However, significant progress has been made since the paper's publication in 2021. LLMs have improved their capabilities, and it is questionable if they can still be called “stochastic parrots.” Instead, they might be seen as black boxes that can reason within a certain context. Our research demonstrates the utility of LLMs in one such context of reasoning. Moreover, several other studies have confirmed the abil of LLMs to reason in other specific contexts~\cite{ahn2024large, mirchandani2023large}. On the other hand, other ---more sensitive--- contexts such as legal or moral reasoning \cite{ALMEIDA2024104145} need more thoughtful human oversight, since a careless integration of LLMs could potentially have an unpredictably disruptive effect. Thus, we suggest approaching LLMs with caution and letting the experiments speak for themselves. After all, technology can improve rapidly, and it is essential to avoid hasty dismissal or overhyping LLMs as a silver bullet.
    \item \textbf{Are private LLMs the sole providers of superior outcomes?} In the past, it would have been accurate to affirm this, as models such as GPT-4 and Claude-3-Opus were recognized for their top-tier response quality. In fact, our research demonstrates that Claude-3-Opus outperforms its competitors. However, the scene is evolving rapidly. Recent models, such as Cohere's Command-R+, Mistral's Mistral models, and Meta's Llama 3 (which we did not incorporate in our study, because its context limit is very low: $8192$), are delivering results comparable to some of the before-mentioned models (e.g.,~\cite{davis2024prompting, liu2023llm360, chen2024chatgpts}). These models also come with open licenses, although with varying levels of permissiveness. Nonetheless, it is important to note that these models are all products of private entities with substantial financial resources. The prospect of a public entity or small business independently creating an LLM from scratch appears remote, largely due to the computational demands and associated costs. This could potentially pose a risk to the transparency of their design processes~\cite{harandizadeh2024risk}.
    \item \textbf{What are the primary obstacles to adopting the strategy proposed in our research?} We identify two significant hurdles: cost constraints and technical limitations. While leveraging LLMs as software-as-service can alleviate the need for in-house cluster training, a substantial financial burden is incurred by processing large volumes of tokens. Our investigation revealed that even with moderate-sized graph instances of around \num{7000} nodes, we quickly reach the token limit of what the most permissive LLM can handle within its context window (see subsection~\ref{subsection:dataset-restriction}). Having said that, recent studies (see, for example,~\cite{munkhdalai2024leave}) have proposed the possibility of “infinite” context windows. Additionally, Google's Gemini Pro 1.5 model boasts an impressive \num{2800000}-token context window limit~\cite{geminiteam2024gemini}. However, to successfully apply a strategy similar to ours in the context of massive graphs---or, more generally, in the context of large-scale problem instances---these two limitations must be significantly mitigated. Alternatively, a novel prompt strategy could be developed to detect patterns in metrics while reducing token counts, potentially through prompt compression~\cite{jiang2023llmlingua} or relevant node metric filtering. Neither of these alternatives has been investigated in our research.
    \item \textbf{As researchers, what other aspects of LLMs should we be cautious about?} Earlier, we touched on the issue of LLMs' capability for reasoning in certain contexts. However, many argue that they are incapable of reasoning altogether. This discrepancy stems from the ambiguity surrounding the term “reasoning”. In our context, reasoning refers to the ability to infer and identify useful patterns in metrics associated with each node of a graph. In particular, we demonstrated that LLMs do not produce random or uninformative values but rather follow the given instructions. However, if we had left the concept of “reasoning” or “pattern discovery” too vague, our research would be susceptible to misinterpretation. Therefore, we recommend that researchers be aware of the underlying philosophical discussions surrounding this technology when working with LLMs. After all, technological advancements can shape the way we express ourselves. A good starting point may be to engage with the works of Floridi (e.g.,~\cite{Floridi2023-FLOAAA-4, Floridi2024-jv}). Particularly since the boundaries of LLMs' capabilities remain undefined and are a subject of ongoing debate.
    \item \textbf{Are there ways to improve the integration between MHs and LLMs?} As discussed in Section~\ref{section:llm}, there are already a few hybridization approaches. These include creating new MHs by leveraging LLMs to generate code and employing LLMs as solvers for optimization problems described in natural language. In contrast, our approach utilizes LLMs as pattern detectors for complex instances. These patterns are then used to bias the search process of the metaheuristic. However, an intriguing integration could involve combining all these hybridization techniques within a single software framework. We believe that these approaches are not mutually exclusive but rather complementary. By unifying them, we may unlock even greater contributions to the field of MHs. One potential approach to achieving this integration could be through the use of agents~\cite{xi2023rise}, which would be responsible for orchestrating the three hybridization methods. Note that agents are currently a topic of significant interest in the LLM community~\cite{guo2024large}.
\end{itemize}

\section{Conclusion}\label{sec:conclusion}

This paper showcased the potential of leveraging Large Language Models (LLMs) as \textit{pattern search engines} to enhance metaheuristics (MH) by integrating the information they provide. We demonstrate the effectiveness of this approach on a combinatorial optimization problem in the realm of social networks. An important aspect of our work is prompt engineering. In fact, useful LLM answers are only obtained with well-designed prompts. In the context of the considered social networks problem, the LLM output is used to compute a probability for each node of the input graph to belong to an optimal solution. These node probabilities are then used to bias the search process of a biased random key genetic algorithm (BRKGA). We could show that our hybrid approach outperforms both the pure BRKGA and the state-of-the-art BRKGA variant, whose search process is biased by the output of a hand-designed (and trained) graph neural network model. To address this, we created a tool named \texttt{OptiPattern} (\textit{LLM-Powered Pattern Recognition for Combinatorial Optimization}), which implements this hybrid method and is available at: \url{https://github.com/camilochs/optipattern}.

This pioneering approach paves the way for further exploration, including extending LLM-assisted pattern recognition to problem instances in a broader range of optimization problems.


\section*{Acknowledgment}

The research presented in this paper was supported by grants TED2021-129319B-I00 and PID2022-136787NB-I00 funded by MCIN/AEI/10.13039/ 501100011033.

\bibliographystyle{IEEEtran}
\bibliography{access}

\begin{IEEEbiography}[{\includegraphics[width=1in,height=1.25in,clip,keepaspectratio]{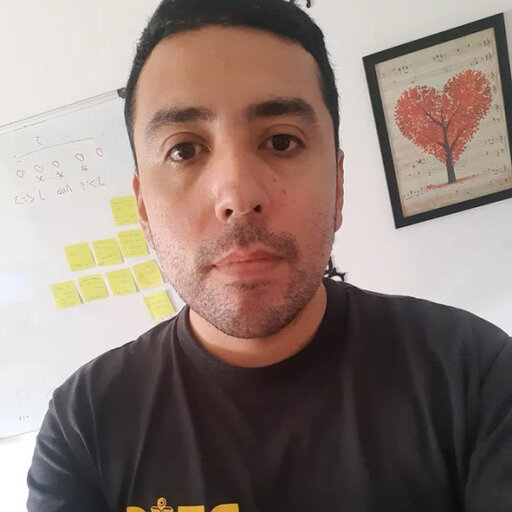}}]{Camilo Chacón Sartori} is currently a PhD student in AI at the Artificial Intelligence Research Institute (IIIA-CSIC) in Bellaterra, Spain. His research focuses on establishing a connection between computational optimization, metaheuristics, visualization tools for understanding algorithm behavior, and generative models.
\end{IEEEbiography}

\begin{IEEEbiography}[{\includegraphics[width=1in,height=1.25in,clip,keepaspectratio]{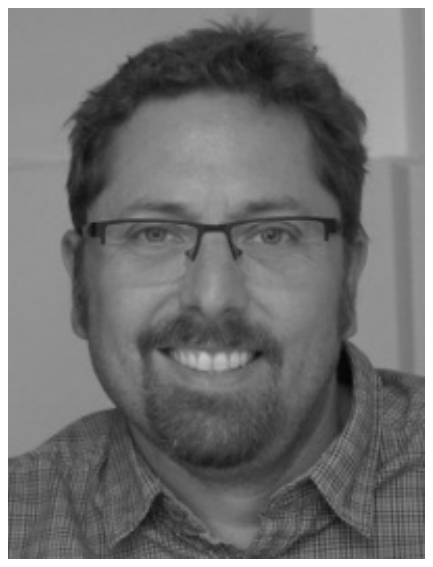}}]{Christian Blum} received the Ph.D. degree in applied sciences from the Free University of Brussels, Brussels, Belgium, in 2004. He is currently a Senior Research Scientist with the Artificial Intelligence Research Institute (IIIA-CSIC), Bellaterra, Spain. His research interests include solving difficult optimization problems using swarm intelligence techniques as well as combinations of metaheuristics with exact techniques.
\end{IEEEbiography}

\begin{IEEEbiography}[{\includegraphics[width=1in,height=1.25in,clip,keepaspectratio]{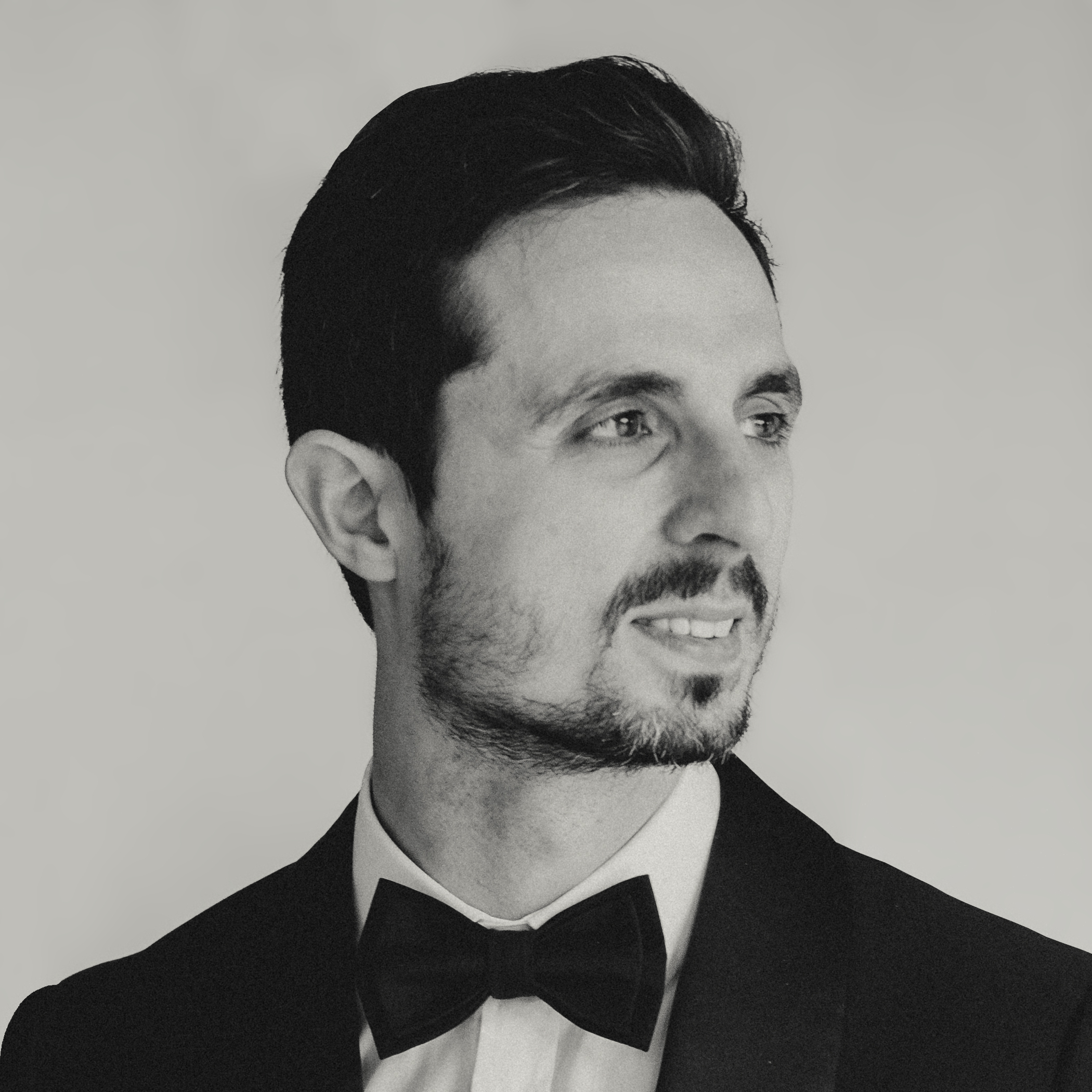}}]{Filippo Bistaffa} received the Ph.D. in Computer Science from the University of Verona in 2016. He is currently a Tenured Researcher (former Marie Skłodowska-Curie Fellow) at the Artificial Intelligence Research Institute (IIIA-CSIC), Bellaterra, Spain. His research interests include optimisation applied to complex real-world problems (e.g., sustainable mobility, team formation for cooperative learning) and, more recently, ethical and trustworthy AI.
\end{IEEEbiography}

\begin{IEEEbiography}[{\includegraphics[width=1in,height=1.25in,clip,keepaspectratio]{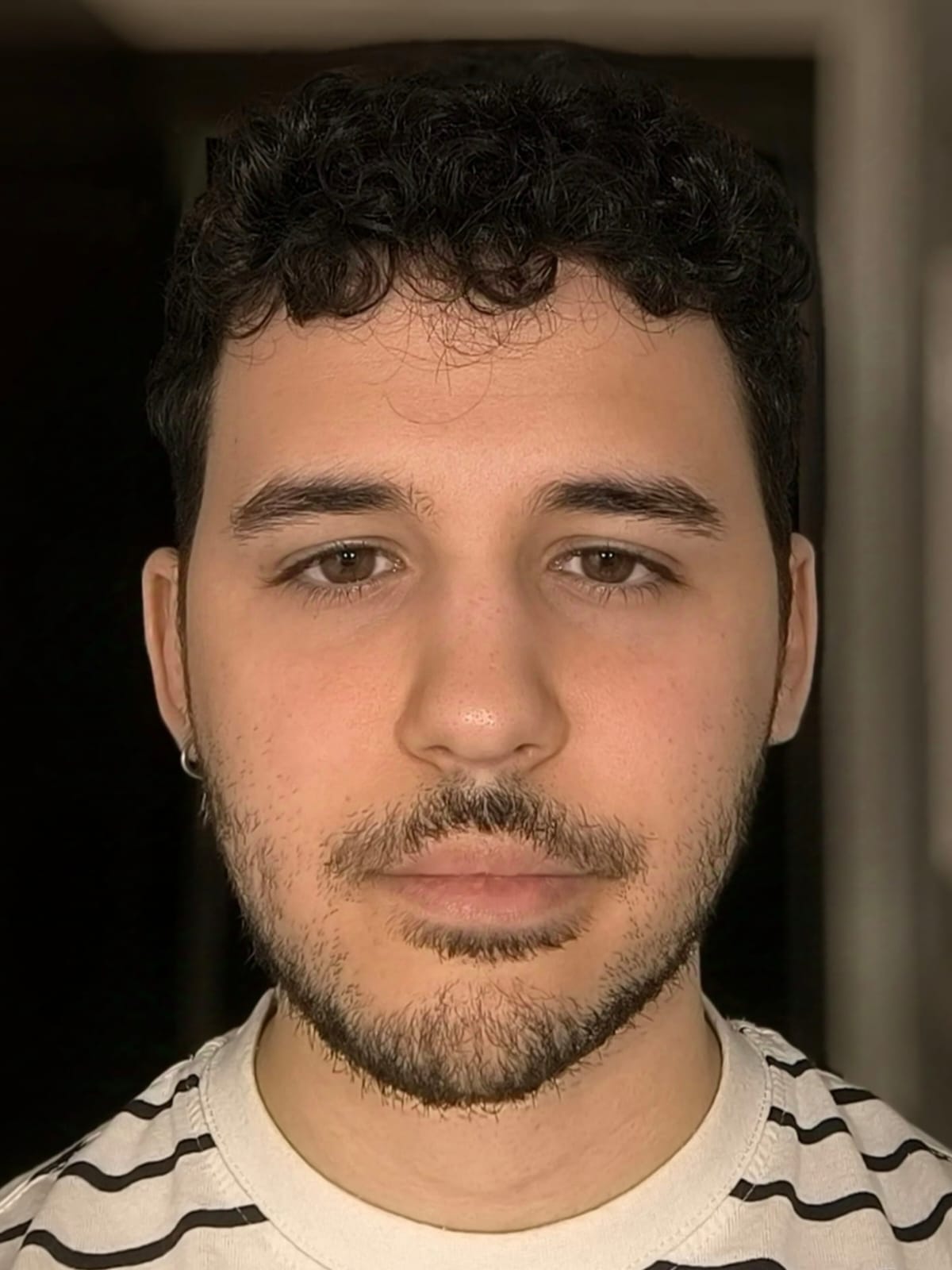}}]{Guillem Rodríguez Corominas} is currently pursuing a PhD in Computer Science at the Polytechnic University of Catalonia (UPC) in Barcelona, Spain, and the Artificial Intelligence Research Institute (IIIA) in Bellaterra, Spain. His research focuses on tackling complex combinatorial optimization problems using metaheuristics, with a special focus on hybridizing these methods with exact algorithms and machine learning techniques.
\end{IEEEbiography}

\EOD

\end{document}